\PassOptionsToPackage{table,xcdraw}{xcolor}

\documentclass[conference]{IEEEtran}

\usepackage[textsize=tiny,colorinlistoftodos, textwidth=1.23cm]{todonotes}

\usepackage{bbm}
\usepackage[ruled,vlined,linesnumbered]{algorithm2e}
\usepackage{amsmath}
\usepackage{multirow}
\usepackage{enumitem}

\usepackage{subcaption}
\usepackage[normalem]{ulem}
\useunder{\uline}{\ul}{}

\usepackage{hyperref}
\hypersetup{
    colorlinks=False,
    linkcolor=blue,
    filecolor=magenta,      
    urlcolor=cyan,
    pdftitle={Overleaf Example},
    pdfpagemode=FullScreen,
    }

\ifCLASSINFOpdf
\else
\fi
\begin{document}
\title{Learning Incident Prediction Models Over Large Geographical Areas for Emergency Response Systems}

\author{\IEEEauthorblockN{Sayyed~Mohsen~Vazirizade}
\IEEEauthorblockA{Vanderbilt University, Nashville, TN\\
   s.m.vazirizade@vanderbilt.edu}
\and
\IEEEauthorblockN{Ayan~Mukhopadhyay}
\IEEEauthorblockA{Vanderbilt University, Nashville, TN\\
    ayan.mukhopadhyay@vanderbilt.edu}
\and
\IEEEauthorblockN{Geoffrey~Pettet}
\IEEEauthorblockA{Vanderbilt University, Nashville, TN\\
    geoffrey.a.pettet@vanderbilt.edu}
\and
\IEEEauthorblockN{Said~El~Said}
\IEEEauthorblockA{Tennessee Department of Transportation\\
   said.elsaid@tn.gov}
\and
\IEEEauthorblockN{Hiba Baroud}
\IEEEauthorblockA{Vanderbilt University, Nashville, TN\\
    hiba.baroud@vanderbilt.edu}
\and
\IEEEauthorblockN{Abhishek~Dubey}
\IEEEauthorblockA{Vanderbilt University, Nashville, TN\\
    abhishek.dubey@vanderbilt.edu}}

\maketitle

\begin{abstract}
Principled decision making in emergency response management necessitates the use of statistical models that predict the spatial-temporal likelihood of incident occurrence. These statistical models are then used for proactive stationing which allocates first responders across the spatial area in order to reduce overall response time. Traditional methods that simply aggregate past incidents over space and time fail to make useful short-term predictions when the spatial region is large and focused on fine-grained spatial entities like interstate highway networks. This is partially due to the sparsity of incidents with respect to the area in consideration. Further, accidents are affected by several covariates, and collecting, cleaning, and managing multiple streams of data from various sources is challenging for large spatial areas. In this paper, we highlight how this problem is being solved for the state of Tennessee,  a state in the USA with a total area of over 100,000 sq. km. 
%Working with the Tennessee Department of Transportation (TDOT) we have developed a novel pipeline to forecast and respond to road accidents on the interstate networks. 
Our pipeline, based on a combination of synthetic resampling, non-spatial clustering, and learning from data can efficiently forecast the spatial and temporal dynamics of accident occurrence, even under sparse conditions. 
In the paper, we describe our pipeline that uses data related to roadway geometry, weather, historical accidents, and real-time traffic congestion to aid accident forecasting. To understand how our forecasting model can affect allocation and dispatch, we improve upon a classical resource allocation approach. Experimental results show that our approach can significantly reduce response times in the field in comparison with current approaches followed by first responders.
\end{abstract}


% \fi
\IEEEpeerreviewmaketitle

\pagenumbering{arabic}
\pagestyle{plain}
\section{Introduction}
% importance of the research \~
% how it is done today (state of the art) \???
% what we wanna do (method) \~
% what are the key results \?

%importance of ERM
A constant threat that plagues humans across the globe are incidents like traffic accidents, fires, and crimes. 
%These concerns are further escalated by rise in population density around the world~\cite{nolan2004establishing}.
Such incidents result in loss of life, injuries, and damage to properties and are collectively labeled as \textit{emergencies}, which are defined as incidents that threaten public safety, health, and welfare. Consider road accidents and calls for emergency medical services (EMS) as examples. Road accidents alone account for 1.25 million deaths globally and about 240 million EMS calls are made in the U.S. each year~\cite{mukhopadhyay2020review}. The large number of such incidents makes it imperative that principled methods be designed to ensure fast and effective response to incidents. At the same time, it is crucial to design infrastructure that mitigates and prevents the occurrence of such incidents. Indeed, it is well-documented that one of the most important responsibilities of federal, state, and local governments is mitigating and dealing with such events~\cite{homeland}. As a result, governments strive to make systematic plans, allocate resources, and take preventive measures in order to alleviate threats that such incidents pose.

%importance of prediction
Emergency response management (ERM) is defined as the set of procedures and tools that first responders use to deal with incidents such as road accidents. It includes specific mechanisms to forecast incidents, detect incidents, allocate resources like ambulances, dispatch resources, and finally mitigate the post-effects of incidents~\cite{mukhopadhyay2020review}. Arguably, the most important component of the pipeline is to understand the spatial and temporal dynamics of incident occurrence. Gaining such an understanding can aid resource allocation and dispatch, improve the understanding of factors that cause accidents, and improve the design of safety codes. While there are several ways to analyze spatial temporal incidents, learning data-driven forecasting models is particularly important since the fundamental goal of understanding the dynamics of accident occurrence is to aid response and dispatch. As a result, \textit{generative} models conditional on relevant covariates are particularly relevant to the overall ERM pipeline. For example, consider a forecasting model for accident occurrence as a function of roadway speed limits. Understanding how the speed limit affects accidents helps in accurately capturing first-order effects that impact accidents, and forecasting future incidents helps shape better policy decisions pertaining to resource allocation (ambulances, for example) and response. 

% of the highways, studying the risk factors, etc. Considering about 1.2 million deaths annually from road accidents alone \cite{roadStats}, analyzing spatial temporal distribution becomes a very crucial problem. \citeauthor{mukhopadhyay2020review}~\cite{mukhopadhyay2020review} and \citeauthor{Lord2010}~\cite{Lord2010} reviewed and summarized various methods and approaches to tackle this problem. 

%goal of the paper and ephasazing on the real application of the research
This paper discusses a framework for predicting extremely sparse spatial temporal incidents. Over the past year, our project has focused on developing principled approaches to address emergency response for Tennessee, a state in the United States with a population of approximately 6.9 million and a total area of over 100,000 sq. km. While we have extensively tackled emergency response in past collaborations with several government bodies restricted to cities~\cite{mukhopadhyayGameSec16,mukhopadhyayAAMAS17,MukhopadhyayICCPS,pettet2020algorithmic,pettet2020hierarchical}, planning emergency response in extremely large geographical areas (an entire state, for example) is significantly more challenging. The problems are exacerbated when we limit the area of interest only to interstate highways across the state, which reduces the number of samples of positive incidents available across the road network, leading to extreme sparsity. This imbalance is particularly evident while creating forecasting models in high temporal resolution. However, our collaboration with first responders revealed that such forecasting models can be extremely beneficial for resource allocation and dispatch.

Our contributions in this paper are two-fold: 1) we develop a pipeline that can effectively forecast incidents that are sparsely scattered in space and time, which can be used by ERM pipelines to reduce the average response time to accidents. We show how a combination of synthetic resampling and non-spatial clustering can result in the creation of accurate spatial temporal models for short-term forecasting of road accidents. 2) Unlike the majority of forecasting models in literature, we evaluate how our forecasting pipeline affects response times to accidents. To this end, we modify the classical p-median problem for resource allocation. Our novel contribution balances the geographic spread of responders and their availability. Through extensive simulations, we show how our forecasting pipeline and allocation algorithm provide significant reduction in response times.

\section{Prior Work}
\label{sec:prior}
A variety of approaches have been used to understand the spatial and temporal dynamics of road accidents. One of the earliest methods, known as `crash frequency analysis', uses the frequency of incidents in a specific discretized spatial area as a measure of the inherent risk the area possesses~\cite{deacon1974identification}. This approach also forms the basis of \textit{hotspot} analysis~\cite{cheng2005experimental,nij2005}, which is widely used in practice as a tool to visualize historical accidents and make predictions. Statistical models have also been explored in this context. The most widely used approach, Poisson regression, models the expected value of the count of incidents in a given time period as a linear combination of the features. While it does not perform well on data with dispersion (where mean is not equal to the variance of the data) and sparse data, hierarchical Poisson models~\cite{Deublein2013, Quddus2008, akinCrashBinomial, ladron2004forecasting, ackaah2011crash} and zero-inflated models can be used instead~\cite{qin2004selecting,lord2007further,huang2010modeling,lord2005poisson}. In recent years, data mining models such as neural networks~\cite{Pande2006,Abdelwahab2002,Chang2005,Riviere2006,zhu2018use,Bao2019} and support vector machines~\cite{doi:10.3141/2024-11,Li2008,Yu2013} have also been explored. We refer interested readers to our survey on emergency response for a detailed analysis of prior work in forecasting, allocation, and dispatch~\cite{mukhopadhyay2020review}.

Two major shortcomings of prior approaches are --- 1) the inability to deal with large sparsity in data, and 2) the inability to make accurate short-term predictions. Even zero-inflated models, the only class of statistical models that have been shown to work fairly well on sparse spatial temporal data, fail to perform well when trained on emergency incidents on highways in large geographic areas (large areas often exhibit $>99\%$ sparsity). We show later how zero-inflated models fail to work in such scenarios. Also, the majority of prior work in accident prediction either focuses on the spatial dynamics of incident occurrence or consider extremely coarse temporal resolutions. While long-term predictions can be useful to analyze policies (optimize road construction, for example), such predictions are not particularly useful for allocating and dispatching responders. We address these gaps in prior work by designing a pipeline for spatial temporal incident prediction that can adapt to extremely sparse data and aid resource allocation. 
\section{Problem Formulation}
\label{sec:problem}

\begin{figure}[t]
\captionsetup{font=small}
    \centering
    \includegraphics[width=\columnwidth]{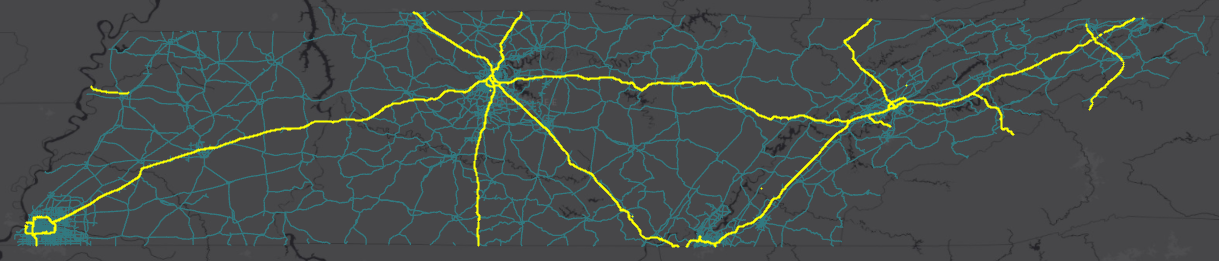}
    \caption{Blue lines represent TN's roadway network. Yellow segments represent interstate highway segments under the jurisdiction of TDOT and are the area of study for this paper.}
    \label{fig:roadsegmenttdot}
\end{figure}

Consider a spatial area of interest $S$, in which incidents (like accidents) occur in space and time. The decision-maker observes a set of samples (possibly noisy) drawn from an incident arrival distribution. These samples are denoted by
% which we refer to as $D$, and can be represented as a vector of tuples
$\{(s_1,t_1,k_1,w_1),(s_2,t_2,k_2,w_2),\\\dots,(s_n,t_n,k_n,w_n)\}$, where $s_i$, $t_i$ and $k_i$ denote the location, time of occurrence, and reported severity of the $i$th incident, respectively, and $w_i \in \mathbbm{R}^m$ represents a vector of features associated with the environment defined by the location and time of the incident. We refer to this tuple of vectors as $D$, which denotes the input data that the decision-maker has access to. The vector $w$ can contain spatial, temporal, or spatio-temporal features and it captures covariates that potentially affect incident occurrence. For example, $w$ typically includes features such as weather, traffic volume, and time of day. The most general form of incident prediction can then be stated as learning the parameters $\theta$ of a function over a random variable $X$ conditioned on $w$. We denote this function by $f(X \mid w,\theta)$. The random variable $X$ represents a measure of incident occurrence such as a \textit{count} or \textit{presence} of incidents during a specific time period. The goal of the incident prediction problem is to find the \textit{optimal} parameters $\theta^*$ that best describe $D$. This can be formulated as a maximum likelihood estimation (MLE) problem or an equivalent empirical risk minimization (ERM) problem.

In our problem setting (the Interstate Highway network of the state of Tennessee) (Fig. \ref{fig:roadsegmenttdot}), it is intuitive to represent $S$ in the form of a graph $G = (V,E)$, where $V$ is a set of vertices and $E$ is a set of edges. Edges represent specific segments of highways. 
% We describe how long highways are divided into smaller segments later. 
% \gp{I don't see discussion on segment division in updated text; am I missing it? SMV: youre right Geof, we dont break the highways into smaller segment, We use inrix segmentation, which is very high resolution. Then we group the neighboring segments to reduce the size of the spatial resolution by about a factor of 3. GAP: Okay, so should we remove this?}
The prediction problem then reduces to learning $f$ such that it captures the spatial temporal dynamics of incidents on $E$.

\subsection{Challenges}

The problem described in previous section is hard due to the following challenges.
%We highlight the following major challenges in learning forecasting models over spatial-temporal incident occurrence. 

\begin{figure}[t]
\captionsetup{font=small}
\centering
\includegraphics[angle =270 ,width=0.8\columnwidth]{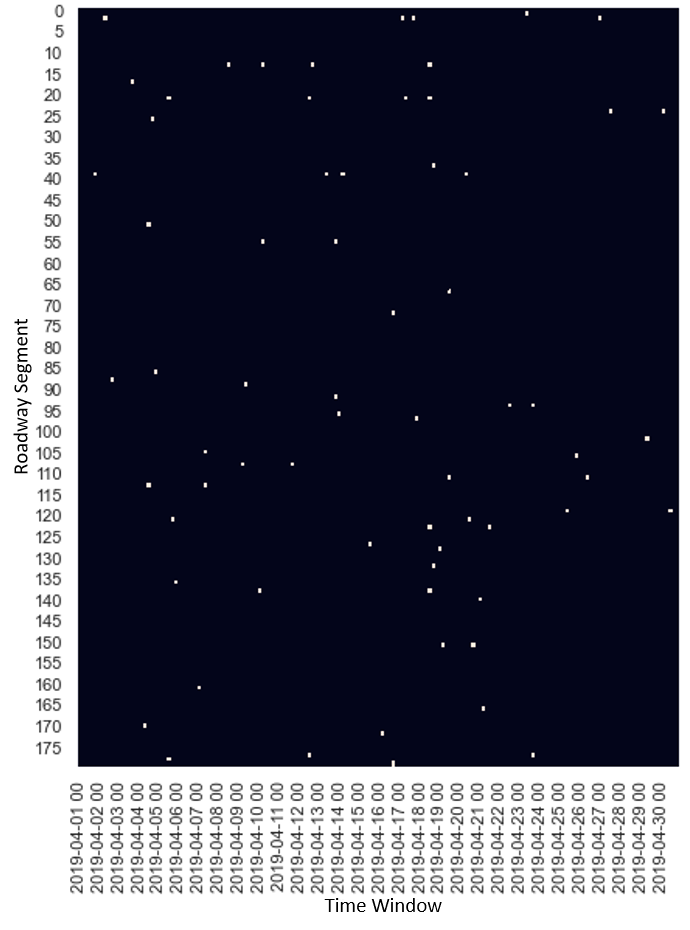}
\caption{Schematic overview of the sparsity of accident occurrence across space and time. The figure shows randomly selected 180 road segments for four hour time windows in April 2019.}
\label{fig:schematic_sparisty_2}
\end{figure}

\begin{enumerate}[leftmargin=* ,noitemsep]
\item \textit{Irregular incident occurrence:} It is well-established in literature that predicting road accidents is extremely difficult due to inherent randomness of accidents and spatially varying factors~\cite{qi2007freeway, mukhopadhyay2020review}.  While accidents are affected by various features, it is difficult to take all determinants into account while designing forecasting models. For example, consider the condition of a specific road. It is difficult to observe such features in real-time, thereby resulting in unobserved heterogeneity in the likelihood of incident occurrence across a large spatial region. Indeed, sophisticated models have under-performed in predicting accidents. For example, an approach particularly important to practitioners was developed by Bao et al.~\cite{Bao2019}, who used a spatio-temporal convolution long short-term memory network (LSTM) to predict short-term crash risks. While the network structure was a combination of various complex sub-networks, the accuracy of hourly predictions was limited, highlighting the inherent difficulty of predicting crash frequency at low temporal and spatial resolutions. 
% \textcolor{red}{ad:but we are also just doing 4 hour predictions...}

\item \textit{Sparsity:} It is also crucial to take into account the frequency of incident occurrence. While the frequency of road accidents is alarming, incidents are actually extremely sporadic when viewed from the perspective of total time and space in consideration. For example, there were a total of approximately 78,000 road accidents reported between 2017-2020 on interstate highways in Tennessee. Now consider the goal of learning the dynamics of incident occurrence. While historical data can be studied using \textit{hotspots} to improve policy, short-term forecasting models are important for deploying ambulances, help-trucks, and other emergency responders. Based on our conversations with first-responders, short term deployment often occurs several times in a day, the most common frequency being once every four hours. Considering a total of about 5,000 road segments and time slots of four hours, our data shows $>99\%$ sparsity. We represent this challenge schematically in Fig. \ref{fig:schematic_sparisty_2} by randomly selecting 180 road segments for April 2019 and 180 four-hour time slots. Each pixel in the matrix denotes the presence (white) or absence (black) of an accident in a segment (denoted by rows) in a span of four hours (denoted by columns). We see that most of the matrix consists of black pixels ($99.8\%$), making such problems extremely difficult from the perspective of data-driven modeling. In comparison, we have previously shown how standard statistical models can be used to predict incidents in small urban areas~\cite{mukhopadhyayGameSec16,mukhopadhyayAAMAS17,MukhopadhyayICCPS} (such situations typically exhibit $<90\%$ sparsity).

\item \textit{Data Integration:} Road accidents are affected by a large number of determinants which can be spatial, temporal, or spatial temporal in nature. For example, the geometry of a specific road segment does not change over time and is an example of a spatial feature. Time of day, on the other hand, is an example of a temporal feature. Some features can be affected by both space and time; for example traffic congestion in a specific area is determined by the spatial location of the area as well as time of day. For predicting accidents in large geographic areas, it is challenging to collect, clean, understand, and analyze data from different sources and integrate them into models for incident prediction.
\end{enumerate}

%We show how we address each of the aforementioned challenges in the next section.

% \begin{table}[t]
% \centering
% \caption{Base Dataset}
% \resizebox{0.48\textwidth}{!}{%
% \begin{tabular}{@{}lllll@{}}
% \toprule
% Name             & Source     & Time Range                  & Size  & Number of Rows \\ \midrule
% Incident Dataset & TDOT       & Feb/01/2017 - May/01/2020   & 21MB  & 80,000         \\
% Weather Dataset  & Weatherbit & Feb/01/2017 - Jun/01/2020   & 300MB & 1,400,000      \\
% Traffic Dataset  & INRIX      & April/01/2017 - Dec/01/2020 & 1.2TB & 30,000,000,000 \\
% Roadway Dataset  & INRIX      & Static                      & 81MB  & 80,000         \\ \bottomrule
% \end{tabular}
% }
% \label{table:base}
% \end{table}

\section{Data}
\label{Data Preparation}

\begin{table*}[]
\centering
\caption{Data Features, Size and Sources}
\resizebox{\textwidth}{!}{%
\begin{tabular}{|l|l|l|l|l|l|l|l|l|}
\hline
\textbf{Dataset}          & \textbf{Range} & \textbf{Size}          & \textbf{Rows}                   & \textbf{Features}                 & \textbf{Source} & \textbf{Frequency} & \textbf{Type}   & \textbf{Description}                                                                                         \\ \hline
-                         &    -            &   -                     &                   -              & Time of day                       & derived         & -                  & Temporal        & We divide each day into six 4-hour time windows.                                                                 \\ \hline
-                         &        -        &    -                    &                          -       & Weekend                           & derived         & -                  & temporal        & A binary feature that denotes weekdays.                                                                      \\ \hline
\multirow{4}{*}{Incident} & 02/01/2017      & \multirow{4}{*}{21MB}  & \multirow{4}{*}{80,000}         & Past Incidents in the last window & derived         & -                  & Spatio-temporal & Number of incidents on the segment in the last time window of 4 hours                                        \\ \cline{5-9} 
                          & to                        &                        &                                 & Past Incidents in a day           & derived         & -                  & Spatio-temporal & Number of incidents on the segment in the last day                                                           \\ \cline{5-9} 
                          & 05/01/2020                       &                        &                                 & Past Incidents in a week          & derived         & -                  & Spatio-temporal & Number of incidents on the segment in the last week                                                          \\\cline{5-9} 
                          &                        &                        &                                 & Past Incidents in a month         & derived         & -                  & Spatio-temporal & Number of incidents on the segment in the last month                                                         \\ \hline
\multirow{4}{*}{Weather}  & 02/01/2017     & \multirow{4}{*}{300MB} & \multirow{4}{*}{1,400,000}      & Visibility                        & Weatherbit      & 1 hour             & Spatio-temporal & A measure of the distance at which an object or light can be clearly discerned.                              \\  \cline{5-9} 
                          & to            &                        &                                 & Wind Speed                        & Weatherbit      & 1 hour             & Spatio-temporal & Speed of wind.                                                                                               \\  \cline{5-9} 
                          & 06/01/2020    &                        &                                 & Precipitation                     & Weatherbit      & 1 hour             & Spatio-temporal & Amount of precipitation.                                                                                     \\  \cline{5-9} 
                          &                &                        &                                 & Temperature                       & Weatherbit      & 1 hour             & Spatio-temporal & It is the reported temperature.                                                                              \\ \hline
\multirow{3}{*}{Traffic}  & 04/01/2017     & \multirow{3}{*}{1.2TB} & \multirow{3}{*}{30,000,000,000} & Congestion                        & derived         & 5 minutes          & Spatio-temporal & Congestion is the ratio of the   difference between free flow speed and the current speed to free flow speed \\  \cline{5-9} 
                          & to             &                        &                                 & Free Flow Speed                   & INRIX           & 5 minutes          & spatial         & The speed at which drivers feel comfortable if there is no traffic and adverse   weather condition.          \\  \cline{5-9} 
                          & 12/01/2020     &                        &                                 & Traffic Confidence                & INRIX           & 5 minutes          & Spatio-temporal & A confidence score regarding the accuracy of the traffic data (we collect this directly from INRIX).         \\ \hline
\multirow{3}{*}{Roadways} & \multirow{3}{*}{Static}         & \multirow{3}{*}{81MB}  & \multirow{3}{*}{80,000}         & Lanes                             & INRIX           & static             & Spatial         & Number of lanes for a roadway segment.                                                                       \\  \cline{5-9} 
                          &                &                        &                                 & Miles                             & derived         & static             & Spatial         & Length of a roadway   segment.                                                                               \\  \cline{5-9} 
                          &                &                        &                                 & iSF                               & derived         & static             & Spatial         & Inverse scale factor which represents the the curvature of a roadway segment.                                \\ \hline
\end{tabular}
}
\label{table:features}
\label{table:base}
\end{table*}

Before describing our pipeline for learning the spatial and temporal dynamics of highway accidents, we describe the covariates used and their sources (Table~\ref{table:base}). We highlight the importance of this stage in real-world machine learning pipelines; in fact, the availability of multiple streams of data has been noted as being particularly important for predicting accidents~\cite{mukhopadhyay2020review}. 
% We also point out that while it might be relatively straightforward to collect, clean, and maintain datasets for transportation problems pertaining to small urban areas, it is significantly harder to do so for entire states. 
Our exercise of collecting data for features that affect accidents was guided by the invaluable domain expertise of first responders and our collaborators at the Tennessee Department of Transportation. 

\subsection{Features}

This section describes the features we extract from the base data (Table~\ref{table:base}) and use as covariates in our pipeline (Table~\ref{Resource Demand Models}).
%in this section.
%We use the following datasets for predictive modeling. For each one, we describe data source, size, time range, granularity, and processing below.

% \usepackage{multirow}

\begin{enumerate}[leftmargin=* ,noitemsep]
\item \textit{Roadway Information}: To learn a predictive model for accidents over a graph of roadways, it is imperative to first define the edges and the vertices of the graph. We collect roadway information from INRIX~\cite{inrix}, a private entity that provides location-based data and analytics, such as traffic and parking, to automakers, cities, and road authorities worldwide. We retrieved information for about 80,000 roadway segments in the state of Tennessee, out of which about 5,000 are interstate highway segments. We also retrieved \textit{static} features associated with each segment that are immutable (relatively) over time. For example, for each road segment, we collected the number of lanes, length, and coordinates. 
% The roadway segments extracted from INRIX makes our spatial resolution. 

In order to evaluate how roadway shape affects accidents, we introduce a feature called the \textit{inverse stretch factor} (iSF), that represents the curvature of road segments. We show an example for calculating iSF in Fig.~\ref{fig:ISF}. For the segment in consideration (between points A and B), iSF can be calculated as the length the straight line $\overline{AB}$ divided by that of the curve $\widehat{AB}$). 
% We calculate iSF for each segment in our dataset.
% It is essentially the reciprocal of the stretch factor and takes a value between 0 and 1.

\begin{figure}[t]
\captionsetup{font=small}
\centering
\includegraphics[width=0.3\columnwidth]{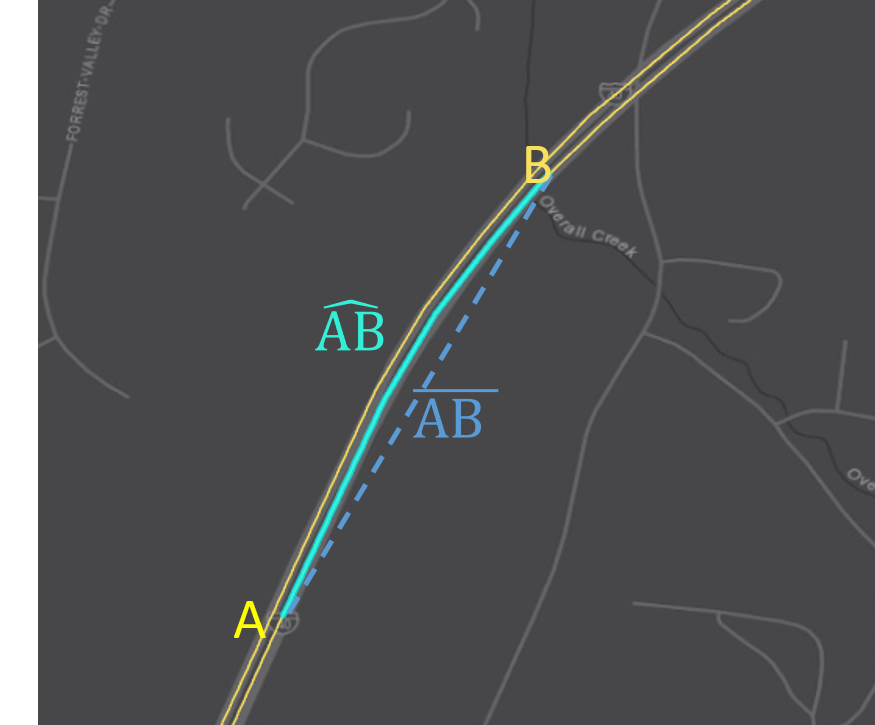}
\caption{A combination of the length of a curve ($\widehat{AB}$) and the shortest path between the two ends of the curve ($\overline{AB}$) can be used to denote its curvature}
\label{fig:ISF}
\end{figure}

\item \textit {Traffic}: The correlation of traffic and road accidents is well-established~\cite{mukhopadhyay2020review}. We collected traffic data for each of the road segments through INRIX at a temporal resolution of 5-minute intervals for about three years. Specifically, we retrieved the free flow speed of traffic, the estimated current speed of the vehicles, and the confidence scores of the estimates. Effective congestion can be calculated from our data as the ratio of the difference between the free flow speed and the current speed to the free flow speed. 
% Due to the the number of the segments and the aforementioned time resolution, the size of our traffic data is 1.2 terabytes, and it has approximately 3e10 rows. 

\item \textit{Weather}: Weather is inherently spatial temporal, and can play an important role in accident rates~\cite{mukhopadhyay2020review}. We collected hourly weather data (temperature, precipitation, visibility, and wind) from 40 different weather stations in and around the state of Tennessee. 
%We show the locations of the stations in Figure~\cref{fig:weather_stations}.
To use weather data to forecast accidents on a given road segment, we use the weather station that is the closest to that particular segment.

% \begin{figure}[t]
% \centering
% \includegraphics[width=3in]{images/weatherbit.png}
% \caption{Location of the weather stations}
% \label{fig:weather_stations}
% \end{figure}
\item \textit{Incidents}: We look at every accident reported in Tennessee from January 2017 to May 2020. Incident data for this project is provided by the Tennessee Department of Transportation (TDOT). Our data consists of approximately 78,000 accidents. The accuracy of the incident data was verified with the Enhanced Tennessee Roadway Information Management System (E-TRIMS). 
% Then, we cleaned and aggregated all available data into the same spatial mapping. 

\end{enumerate}

%\cref{table:features} summarizes the  features we derived from the data (\cref{table:base}). 

% \input{tables/table2}

\begin{figure}[t]
\captionsetup{font=small}
\centering
\begin{center}
\includegraphics[width=\columnwidth]{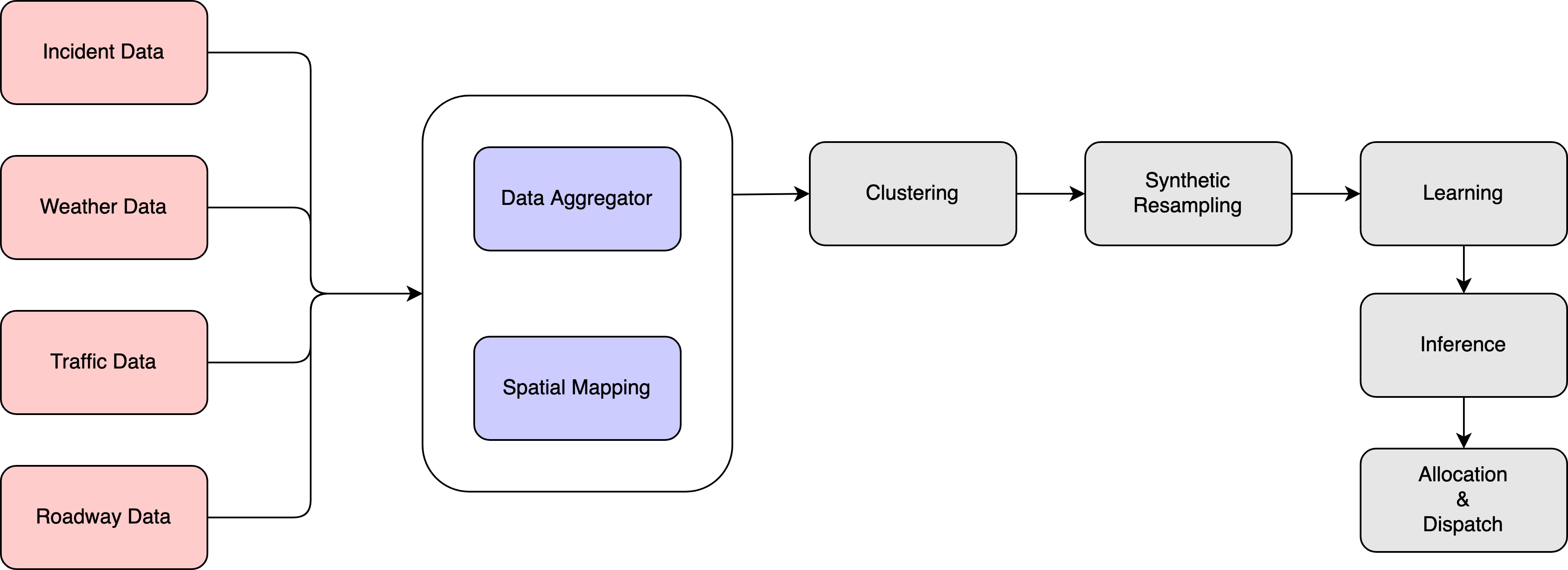}
\caption{Overview of our approach. We extract spatial temporal information from a variety of data sources, focus on heterogeneity not explicitly modeled in the feature space by identifying clusters, perform synthetic sampling to address sparsity, and learn models on incident occurrence. 
%\textcolor{red}{please check the section numbers.}
}
\label{fig:outline}
\end{center}
\end{figure}

\section{Approach}
\label{Resource Demand Models}

We now describe how we design a pipeline to predict roadway accidents in space and time. To begin with, we filter out road segments that exhibit no accidents or extremely small number of accidents over the temporal period in consideration (about three years). Our analysis is done on $77\%$ of the observed accidents with a total sparsity of $98\%$. Recall our goal is to learn a function $f$ (section~\ref{sec:problem}) that outputs the likelihood of incident occurrence on a road segment conditional on a set of features. A straightforward way to do so is to learn a separate model over each segment. However, such an approach results in \textit{overfitting}; each segment contributes a relatively small amount of data which ignores structural similarities between patterns of incident occurrence across the entire spatial region in consideration. The other approach is to learn one model for the entire area. However, a \textit{universal} model fails to capture any heterogeneity that is not explicitly modeled in the feature space. In order to balance these considerations, we try to identify segments that observe similar patterns for incident occurrence~\cite{mukhopadhyayGameSec16,mukhopadhyayAAMAS17}. While it is possible to identify distinct spatial regions (hotspots) and learn a separate model for each area, it is possible that there exists generalizable information in the entire area that is spatially invariant.
In order to do so, we seek to identify common areas irrespective of spatial contiguity by clustering all the available segments based on their frequency of incident occurrence. In this study, we used the well known \textit{k-means} algorithm~\cite{macqueen1967some} to group the segments into distinct clusters.   

Given clusters of roadway segments that share similar patterns of spatial-temporal incident occurrence, learning the patterns is still challenging due to the sparsity of the data. To address this concern we perform synthetic under-sampling and over-sampling to balance our data. However, naive synthetic sampling performs poorly in our case since the relative frequencies of incident occurrence are markedly different among the clusters. Therefore it is impractical to `balance' data in each cluster in the same manner. To alleviate this, we start with the cluster with the highest frequency of incident occurrence (cluster A, say) and perform synthetic sampling such that the number of positive data points (spatial segments in temporal windows that \textit{have} accidents) is the same as the number of negative data points (spatial segments in temporal windows that \textit{do not} have accidents). Then, we perform synthetic sampling in the other clusters such that the ratio of accidents occurring for any given cluster (cluster B, say) to the frequency in A is the same as in the original dataset.\footnote{we also show results without synthetic sampling and clustering.}

Clustering and synthetic sampling provide  the foundation for learning spatial temporal forecasting models over accident occurrence. We use the following well-known models to this end.

\begin{enumerate} [leftmargin=*,noitemsep]
    \item \textbf{Logistic Regression}: There are two classes of approaches that can be used to forecast the chances of accidents on road segments. First, one can try to model the count of accidents as a binary variable and use well-known count-based regression models like Poisson regression, zero-inflated Poisson regression, and negative binomial regression~\cite{mukhopadhyay2020review}. The other approach is to treat the occurrence of accidents as a dichotomous output and model the likelihood that \textit{any} accidents occur. We start with the latter, and use logistic regression, which models the log-odds of the probability of incident occurrence as a linear combination of the features $w$. 
    % While training the logistic regression model, we tune the threshold for classification through cross-validation to maximize the resulting F1-score.
    
    %Among the all regression models, logistic regression is a useful but simple method for binary classification. It is easy to implement, interpret, and very efficient to train. However, it might not be flexible enough to provide accurate results. logistic regression is used to predict the accident occurrence or hotspot prediction. Previous studies used logistic regression as well as  logistic regression combined with other models to solve accident related problems \cite{7232194, kim2007modeling, ALGHAMDI2002729}
    % %First a simple logistic regression model is applied on the clustered data. Using train data the model is trained and using the verification data the threshold is adjusted. Subsequently, in two separate analyses, a simple logistic regression is applied on the clustered and then resampled (using ROS and RUS).
    \item \textbf{Zero-Inflated Poisson Regression}:
    We also use count-based models to model accident occurrence conditional on spatial temporal features. While Poisson regression has been widely used to model accident occurrence~\cite{mukhopadhyay2020review}, hierarchical Poisson models and zero-inflated models have demonstrated significantly improved predictive power~\cite{lord2007further}. Zero-inflated models can be described as having dual states, one of which is the \textit{normal} state, and the other the \textit{zero} state~\cite{li1999multivariate}. 
    % The excess zeros that cannot be explained by standard count-based models can then be considered to have arisen due to the presence of a separate state. 
    % Zero-inflated models have shown better statistical fit to accident data than traditional count based models~\cite{qin2004selecting,lord2007further}.
    % \item \textbf{Negative Binomial Regression}:
    % We also use a hierarchical Poisson model to forecast accident counts. Standard Poisson models assume that the mean of the sample equals its variance. However, accident data has been shown to be over-dispersed (the variance of the data exceeds the mean). Hierarchical models address this issue by modeling the mean of the Poisson distribution as a random variable itself. We use the well-known negative binomial model, which is essentially a Poisson distribution whose mean parameter follows a gamma distribution.
    \item \textbf{Random Forests}:
    Random forest classifiers are a decision tree ensemble method where each tree is constructed from independently bootstrapped samples \cite{breiman2001random}. They reduce model variance and are less likely to overfit compared to standard decision trees due to bootstrap aggregation and the use of a random selection of features to split nodes when constructing each tree (called ``feature bagging''). In addition to synthetic sampling, random forests can address sparsity using the Balanced Random Forest method \cite{chen2004imbalRF}. This works by assigning weights to each class inverse-proportionally to their frequency in the dataset, giving a heavier penalty to misclassifying the minority class. 
    \item \textbf{Artificial Neural Networks}:
    Finally, we also use simple artificial neural networks to learn a model over incident occurrence. Neural networks consist of a set of \textit{layers}, each of which further consists of \textit{neurons} or computing units. The output of each layer is fed as input to the next layer~\cite{Goodfellow-et-al-2016}. Each neuron uses a non-linear function (called the activation function) of the sum of its inputs, and produces an output. The network can be trained by stochastic gradient descent. We use fully connected layers in this study. An important note to practitioners is the non-interpretability of neural networks can be a barrier when deploying systems in the real-world that affect government policies.
    % with the rectified linear unit (relu) as the activation function. 
    % We use a single neuron at the last layer with a sigmoid activation to limit the output of the entire network between 0 and 1. 
    % We tune the number of layers, number of neurons in each layer, and classification threshold of the network through cross-validation. We point out that traditionally, complex neural networks have been shown to perform poorly with regard to sparse spatial temporal predictions~\cite{Bao2019,mukhopadhyay2020review}. Further, 
\end{enumerate}

It is natural to compare forecasting approaches through metrics like likelihood values on test data, error rates, precision, and recall. However, our conversations with first responders revealed that it is particularly beneficial for them to understand if forecasting models can \textit{rank} roadway segments based on risk. This is intuitive since accurately forecasting the risk at each segment relative to other segments is important for allocating resources. Therefore, besides standard statistical metrics (accuracy, precision, recall, F1-score), we also report the correlation of each model’s marginal accident likelihood distribution over space with the real accident distribution. We report both Pearson and Spearman correlation values.

% We also recommend that model designers use this metric in the absence of access to transportation simulators to evaluate the effect of forecasting on allocation and dispatch.
% \gp{If we recommend designers use correlation, why do we use F1 score when tuning hyperparameters? And have we checked that correlation scores actually ``correlate'' with the distance metric in our allocation?}

\section{Allocation and Dispatch}
\label{sec:allocation}
The primary purpose of incident prediction models is to make informed resource allocation decisions. However, prior literature rarely evaluates ERM pipelines in their entirety. Our goal in this project is to ensure that helper vehicles and ambulances controlled by the state of Tennessee save vital response time when dealing with accidents. Therefore, we seek to evaluate the impact incident models have on response time outcomes. Our evaluation process is guided by the following steps:

\begin{enumerate}[leftmargin=*,noitemsep]
    \item \textbf{Understanding existing policies}: Through our collaboration with first responders, we first understand how emergency resources are allocated and deployed in practice. We describe this approach below and use it as our baseline. 
    \item \textbf{Resource Allocation}: In practice, discrete location models like the well-known p-median formulation~\cite{calvo1973location, serra1998p, dzator2013effective, caccetta2005heuristic} are widely used to allocated emergency resources~\cite{mukhopadhyay2020review}. A shortcoming of such approaches is that the service time of resources (for example, the time that ambulances are busy responding to accidents) is not taken directly into account in the allocation process. 
    % \igp{Several coverage models account for responder unavailability, including MEXCLP and MALP. We shouldn't use this as a justification for p-median.}
    We introduce a novel modification to the p-median problem to design a heuristic approach that bridges this crucial gap.
    \item \textbf{Evaluation}: Using the proposed allocation model, we compare the performance of existing prediction models and our pipeline by creating a black-box simulator that imitates emergency response. 
\end{enumerate}

\subsection{Resource Allocation}
In practice, our interaction with first responders revealed that resource allocation is based on identifying \textit{hotspots} of incident occurrence. First, a map based on historical accidents is created. Then a group of experienced engineers determine the location of the responders; typically, responders are placed in areas with the highest historical accident rates. 

\begin{figure}[h]
\captionsetup{font=small}
\centering
\includegraphics[height=1in]{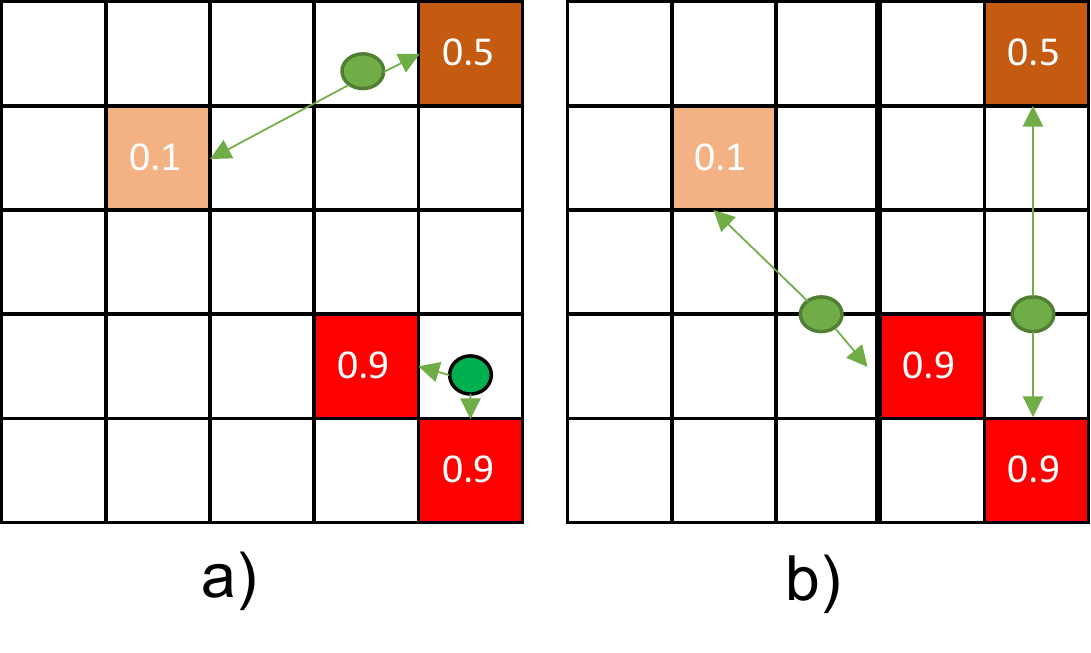}
\caption{Illustrating the impact of $\alpha$: a) standard p-median ($\alpha=0$). b) modified p-median with $\alpha>0$. Notice as $\alpha$ increases responders (green dots) are tightly packed around high demand areas.}
%(Schematic) Effect of choice of $\alpha$ on responder allocations}
\label{fig:alphaAllocation}
\end{figure}

The allocation formulation we use is based on the p-median problem, which is commonly applied to ambulance allocation. The objective of the standard p-median problem is to locate $p$ facilities (i.e. responders) such that the average demand-weighted distance between edges and their nearest facility is minimized. One shortcoming of the p-median formulation is that it does not account for responders becoming unavailable when attending to incidents. To address this, we modify the standard p-median formulation by adding a balancing term to the objective function. Intuitively, this balancing term penalizes responders that cover disproportionately large demand compared to other facilities, encouraging multiple responders to congregate near high demand areas. This effect is schematically demonstrated in Fig.~\ref{fig:alphaAllocation}. In the figure, the values in the cells correspond to the chance of accident occurrence for the location and the green points show the allocated locations of responders ($p$=2 in this case). By considering $\alpha=0$ (alternative a), the problem is equivalent to the simple p-median formulation, which seeks to minimize the weighted distance between allocations and points of demand. However, by increasing $\alpha$ (alternative b), the optimizer seeks to avoid assigning high risk cells to a single responder. Formally, we solve the following optimization problem:

% old formulation
% \begin{subequations}
% \begin{align}
%     &\min_{k} \sum_{i=1}^{i=n} \sum_{j=1}^{j=p} d_{ij} k_{ij} a_{i} b_{j}\\
%     & \text{s.t.} \sum_{i=1}^{i=n} \sum_{j=1}^{j=p} k_{ij} = p
% \end{align}
% \end{subequations}
%
% our cost function is $Cost= \sum_{i=1}^{i=n} \sum_{j=1}^{j=p} d_{ij} k_{ij} a_{i} b_{j}$

% exact formulation
\small
\begin{subequations}
\begin{align}
    \text{min} & \sum_{i=1}^{|E|} \sum_{j=1}^{|L|} a_{i} d_{ij} Y_{ij} b_{j} \label{allocation_objective} \\
    \text{s.t.} &  \sum_{j=1}^{|L|} Y_{ij} = 1, \ \ \ \forall i \in \{1,\dots,|E|\} \label{allocation_meet_demand} \\
    & \sum_{j=1}^{|L|} X_j = p \label{allocation_p} \\
    & Y_{ij} \leq X_j, \ \ \ \forall i \in \{1,\dots,|E|\}, \forall j \in \{1,\dots,|L|\} \label{allocation_placed_responder} \\
    & X_{j}, Y_{ij} \in \{0, 1\}, \ \ \ \forall i \in \{1,\dots,|E|\}, \forall j \in \{1,\dots,|L|\}
\end{align}
\end{subequations}

\normalsize
\noindent where $E$ is the set of demand edges from graph $G$, $L$ is the set of possible responder locations, $p$ is the number of responders to be located, $a_{i}$ is the likelihood of accident occurrence on edge $e_i \in E$, and $d_{ij}$ is the distance between edge $e_i \in E$ and location $l_j \in L$. $Y_{ij}$ and $X_j$ are two sets of decision variables; $X_j = 1$ if a responder is located at $l_j \in L$ and $0$ otherwise, and $Y_{ij} = 1$ if edge $e_i \in E$ is covered by a responder located at $l_j \in L$ (i.e. the responder at $j$ is the nearest placed responder to $e$) and $0$ otherwise. The balancing term we add is denoted by $b_j = {(\frac{\sum_{e \in E} a_{e}Y_{ej}}{\sum_{e \in E} a_{e}})^\alpha}$, and represents the proportion of total demand covered by a responder located at $j$. The influence of the balancing term is controlled by the hyper-parameter $\alpha$; intuitively, as $\alpha$ increases, responders are more `tightly packed' around high demand areas, and if $\alpha = 0$ our formulation reduces to the standard p-median formulation. 
% In other words, it attempts to distribute the demand equally between the responders while minimizing the the weighted distance. By doing so, the two high risk cells are assigned to two different responders.
Constraint (\ref{allocation_meet_demand}) expresses that the demand of each edge must be met, (\ref{allocation_p}) ensures that $p$ responders are located, and (\ref{allocation_placed_responder}) shows that edges must be covered only by locations where responders have been located. 
% \gp{algorithm notation does not match optimization problem (for example, uses $J$ for location set rather than $L$}

% \begin{figure}[t]
% \centering
% \includegraphics[width=3in]{images/temp_TN_grid.png}
% \caption{(Schematic) Grid of possible responder locations across Tennessee, with incident frequency encoded as color. }
% \label{fig:resp_location_grid}
% \end{figure}

\small
\begin{algorithm}[t]
\SetAlgoLined
\SetKwInOut{Input}{input}
\SetKwInOut{Output}{output}
\Input{Demand Edges $E$, Potential Responder Locations $L$, Segment Incident Likelihoods $a_i \ \ \forall e_i \in E$, Segment to Location Distances $d(i, j) \ \ \forall e_i \in E, \forall l_j \in L$, Number of Responders $p$, Balance Factor $\alpha$}
\Output{Responder Locations $X$}

Initialize $k := 0$, $X_k := \emptyset$ \; \label{algo:ga_init}
\While{$k < p$}{
    $k := k + 1$\; \label{algo:ga_iter_update}
    % Compute $Z_j^k := \sum_{i \in I} a_i d(i, j \cup X_{k-1}) \ \ \forall j \in J, \ j \notin X_{k-1}$ \; 
    % Compute $Z_j^k := \sum_{i \in I} a_i d(i, j \cup X_{k-1}) b_{j}$ \ for each location $j$ that is not in $X_{k-1}$\;
    
    \For{$\mathrm{location } \ l_{j'} \in L, \ \ \mathrm{where } \ j' \notin X_{k-1}$}{
        $X'_{k} := X_{k-1} \cup l_{j'}$\; \label{algo:ga_start_score}
        Find nearest facilities $y_i \ \forall e_i \in E$, where $y_{e_i} \in X'_k$\;
        Compute balance terms $b_{j} := {(\frac{\sum_{e_i \in E} a_{i}\psi}{\sum_{e_i \in E} a_{i}})^\alpha} \ \ \forall l_j \in L$ where $\psi := 1$ if $y_i=l_j$, $\psi := 0$ otherwise\;
        Compute $Z^k_{j'} := \sum_{e_i \in E} a_e d(e_i, y_i) b_{y_{e_i}}$\; \label{algo:ga_end_score}
    
    }
    Best location $l_j^* := \text{argmin}_j \ Z^k_j$\; \label{algo:ga_find_best}
    $X_k := X_{k-1} \cup j^*$\; \label{algo:ga_add_best}
}
Return $X_k$
\caption{Greedy-Add Algorithm}
\label{algo:greedy_add}
 %\vspace{-0.2in}
\end{algorithm}
\normalsize
The p-median problem is known to be NP-hard on general networks~\cite{kariv1979algorithmic}, therefore heuristic methods are employed to find approximate solutions in practice. 
% Evaluating the allocation method is not the goal of this paper, so we are not concerned with finding the best heuristic. 
We use the Greedy-Add algorithm~\cite{daskin2011network} to optimize the locations of responders. We show the algorithm in Algorithm \ref{algo:greedy_add}. 
% long-form explanation
First we initialize the iteration counter $k$ and the set of allocated responder locations $X_k$ to the empty set (step \ref{algo:ga_init}). Then, as long as there are responders awaiting allocation, we iterate through the following loop: (1) update counter $k$ current iteration (step \ref{algo:ga_iter_update}), (2) for each potential location not already in the allocation, compute the modified p-median score (equation \ref{allocation_objective}) of the allocation which includes the potential location (steps \ref{algo:ga_start_score} - \ref{algo:ga_end_score}), and (3) find the location that minimizes the modified p-median score (step \ref{algo:ga_find_best}) and add it to the set of allocated responder locations (step \ref{algo:ga_add_best}).
%shorter, intuitive explanation
%In Greedy-Add, facilities are located one at a time, each being placed at the location which minimizes the cost function at that step, until p facilities are located. 
While myopic, this algorithm is scalable to large allocation problems.

Rather than restricting responders to the roadway segments $E$, we allow them to be located anywhere across the state. To accomplish this, we define the set of possible responder locations $L$ as a grid of spatial cells over Tennessee. We define each grid cell as being $0.1$ degrees latitude by $0.1$ degrees longitude, which is approximately 9km x 11km in Tennessee. This results in 1445 possible locations across the state.
%which are shown in \cref{fig:resp_location_grid}. 
The center of each cell is used when calculating the distance between it and each edge in $E$.  

Given an allocation of responders, we simulate response to real incidents to evaluate the efficacy of our model. Response to emergency incidents is typically greedy~\cite{mukhopadhyay2020review}; the closest available responder to the scene of the incident is dispatched to attend to it. This a direct consequence of the critical nature of the incidents that emergency responders address. We use a simulator that imitates greedy dispatch and evaluate the performance of different predictive models.

\section{Experimental Evaluation}
\label{sec:experiments}
To evaluate our models, we use actual historical incident data, roadway geometry, traffic data, and weather data. We train each model based on a rolling temporal window. We start with a train set of 10 months and use the next month as the evaluation set. Then, we use a rolling temporal window and add one month to the training set and use the subsequent month as the test set. Our code, a synthesized dataset, and high-resolution figures and tables are all available online (see the Appendix).

\subsection{Model hyper-parameters} We tune hyper-parameters for each model by cross-validation. For models based on random forests and neural networks, we keep the architecture fixed based on the largest training sample we have; classification thresholds are tuned for every training window based on a validation set. We describe our model parameters below:

\begin{enumerate}[leftmargin=*,noitemsep]
    \item Random forests: Each random forest consists of 250 decision trees. We use Gini impurity to measure the quality of a node split, and consider $\sqrt{\mid w \mid}$ random features for each split, where $w$ is the total number of features. The following hyper-parameters are tuned for each model: the maximum depth of each tree, the minimum number of observations in a node required to split it, and the minimum number of samples required to be at a leaf node to split it's parent. %\gp{Not sure if this should go here or in experiments: number of trees is $250$, number of features to consider when looking for the best split is $\sqrt{num \ features}$}
    \item Neural networks: We use a sequential architecture with fully connected layers. We use a total of three hidden layers. The size of the first layer equals twice the number of input features $w$ (the number of neurons in the input layer). The second and third layers consist of neurons equal to the size of the input layer. The output layer conists of a single neuron. We use the `ReLU' activation function~\cite{agarap2018deep} for all hidden layers and the sigmoid activation function for the output layer. We minimize the cross-entropy loss between true labels and predicted labels and use the \textit{adam} algorithm~\cite{kingma2014adam} for training the network.
    \item Clustering: We use the \textit{k-means} algorithm~\cite{macqueen1967some} to group the segments into clusters. We use $k=2$. A higher value for $k$ renders an extremely small number of segments in some of the clusters, thereby hampering overall performance. Naturally, we recommend practitioners to tune all hyper-parameters based on the specific dataset in consideration.
\end{enumerate}

% \begin{figure}[t]
% \captionsetup{font=small}
% \centering
% \includegraphics[height=1in]{images/traininset.pdf}
% \caption{Each row describes a test case. The months in yellow are the forecasting target. The months in red are used to train the models for prediction for the specific row.}
% %The sequence of the pairs of train datasets and test datasets; 12 pairs in total. In each row, the  months in red are used for training to forecast the incidents in each segment for month in yellow. } %-\textcolor{red}{caption is not very easy to understand} }
% \label{fig:moving_window}
% \end{figure}

% \usepackage{multirow}
\begin{table}[]
\centering
\caption{Summary of performance evaluation metrics for each model in percentage (the performance in each column is color coded; green is the best and red is the worst)}
\resizebox{0.48\textwidth}{!}{%
\begin{tabular}{|l|l|l|l|c|c|c|c|c|c|}
\hline
\textbf{Model}         & \textbf{Clustering}          & \textbf{Resampling} & \textbf{Name} & \textbf{Accuracy}                     & \textbf{Precision}                    & \textbf{Recall}                       & \textbf{F1-Score}                     & \textbf{Pearson}                      & \textbf{Spearman}                     \\ \hline
\multicolumn{3}{|l|}{Naive}                                                 & Naïve         & \cellcolor[HTML]{63BE7B}\textbf{95.5} & \cellcolor[HTML]{F8696B}3.8           & \cellcolor[HTML]{F8696B}4.2           & \cellcolor[HTML]{F8696B}4.0           & \cellcolor[HTML]{63BE7B}\textbf{82.1} & \cellcolor[HTML]{BCE2C8}60.8          \\ \hline
                       &                              & No   resampling     & LR+NoR+NoC1   & \cellcolor[HTML]{FAD1D4}94.0          & \cellcolor[HTML]{FACED0}13.8          & \cellcolor[HTML]{FBE8EB}27.4          & \cellcolor[HTML]{FBDCDF}18.2          & \cellcolor[HTML]{FAD0D3}70.4          & \cellcolor[HTML]{FAC3C5}55.2          \\ \cline{3-10} 
                       &                              & RUS                 & LR+RUS+NoC1   & \cellcolor[HTML]{F99395}93.0          & \cellcolor[HTML]{FAC4C6}12.8          & \cellcolor[HTML]{C1E4CC}32.3          & \cellcolor[HTML]{FBDCDF}18.3          & \cellcolor[HTML]{F87F82}63.1          & \cellcolor[HTML]{FAB8BB}54.7          \\ \cline{3-10} 
                       & \multirow{-3}{*}{No cluster} & ROS                 & LR+ROS+NoC1   & \cellcolor[HTML]{F99496}93.0          & \cellcolor[HTML]{FAC4C7}12.8          & \cellcolor[HTML]{C2E5CD}32.3          & \cellcolor[HTML]{FBDDE0}18.3          & \cellcolor[HTML]{F88083}63.2          & \cellcolor[HTML]{FAB9BC}54.7          \\ \cline{2-10} 
                       &                              & No sample           & LR+NoR+KM2    & \cellcolor[HTML]{F99598}93.0          & \cellcolor[HTML]{FAC1C4}12.5          & \cellcolor[HTML]{FBFBFE}30.9          & \cellcolor[HTML]{FBD8DB}17.7          & \cellcolor[HTML]{CDE9D7}76.6          & \cellcolor[HTML]{F4F9F8}58.4          \\ \cline{3-10} 
                       &                              & RUS                 & LR+RUS+KM2    & \cellcolor[HTML]{F8696B}92.3          & \cellcolor[HTML]{FABCBF}12.1          & \cellcolor[HTML]{63BE7B}\textbf{34.4} & \cellcolor[HTML]{FBD9DC}17.8          & \cellcolor[HTML]{FBFBFE}74.2          & \cellcolor[HTML]{FCFCFF}58.1          \\ \cline{3-10} 
\multirow{-6}{*}{LR}   & \multirow{-3}{*}{clustering} & ROS                 & LR+ROS+KM2    & \cellcolor[HTML]{F86E70}92.4          & \cellcolor[HTML]{FABEC0}12.2          & \cellcolor[HTML]{6FC386}34.2          & \cellcolor[HTML]{FBD9DC}17.9          & \cellcolor[HTML]{FCFCFF}74.2          & \cellcolor[HTML]{FCFCFF}58.1          \\ \hline
                       &                              & No   resampling     & NN+NoR+NoC1   & \cellcolor[HTML]{D5ECDD}94.9          & \cellcolor[HTML]{C2E5CD}19.2          & \cellcolor[HTML]{ADDCBB}32.8          & \cellcolor[HTML]{7DC992}24.0          & \cellcolor[HTML]{FBE0E2}71.7          & \cellcolor[HTML]{F2F8F7}58.5          \\ \cline{3-10} 
                       &                              & RUS                 & NN+RUS+NoC1   & \cellcolor[HTML]{CCE9D6}95.0          & \cellcolor[HTML]{C0E4CB}19.2          & \cellcolor[HTML]{B4DFC1}32.6          & \cellcolor[HTML]{7AC88F}24.1          & \cellcolor[HTML]{FBF0F3}73.2          & \cellcolor[HTML]{E1F1E8}59.3          \\ \cline{3-10} 
                       & \multirow{-3}{*}{No cluster} & ROS                 & NN+ROS+NoC1   & \cellcolor[HTML]{DBEFE3}94.9          & \cellcolor[HTML]{C8E7D3}19.1          & \cellcolor[HTML]{AADBB8}32.8          & \cellcolor[HTML]{83CB97}23.9          & \cellcolor[HTML]{FAC4C7}69.3          & \cellcolor[HTML]{FABABC}54.7          \\ \cline{2-10} 
                       &                              & No sample           & NN+NoR+KM2    & \cellcolor[HTML]{C4E6CF}95.0          & \cellcolor[HTML]{C9E8D3}19.0          & \cellcolor[HTML]{DEF0E5}31.6          & \cellcolor[HTML]{94D2A5}23.7          & \cellcolor[HTML]{E2F2E8}75.6          & \cellcolor[HTML]{E9F5EF}58.9          \\ \cline{3-10} 
                       &                              & RUS                 & NN+RUS+KM2    & \cellcolor[HTML]{FCFCFF}94.7          & \cellcolor[HTML]{F6FAFA}18.4          & \cellcolor[HTML]{B0DDBD}32.7          & \cellcolor[HTML]{ADDCBB}23.3          & \cellcolor[HTML]{FBEFF2}73.1          & \cellcolor[HTML]{FAB7BA}54.6          \\ \cline{3-10} 
\multirow{-6}{*}{NN}   & \multirow{-3}{*}{clustering} & ROS                 & NN+ROS+KM2    & \cellcolor[HTML]{FBF8FB}94.7          & \cellcolor[HTML]{FCFCFF}18.3          & \cellcolor[HTML]{9DD6AD}33.1          & \cellcolor[HTML]{ABDBB9}23.3          & \cellcolor[HTML]{F7FAFB}74.5          & \cellcolor[HTML]{FAC8CA}55.4          \\ \hline
                       &                              & No   resampling     & RF+NoR+NoC1   & \cellcolor[HTML]{C8E7D2}95.0          & \cellcolor[HTML]{CEEAD7}19.0          & \cellcolor[HTML]{D6EDDE}31.8          & \cellcolor[HTML]{96D3A7}23.6          & \cellcolor[HTML]{A5D9B4}78.7          & \cellcolor[HTML]{80CA94}63.4          \\ \cline{3-10} 
                       &                              & RUS                 & RF+RUS+NoC1   & \cellcolor[HTML]{ACDCBA}95.2          & \cellcolor[HTML]{BAE2C6}19.3          & \cellcolor[HTML]{FBF9FC}30.5          & \cellcolor[HTML]{9FD6AE}23.5          & \cellcolor[HTML]{F9AFB2}67.4          & \cellcolor[HTML]{FBE4E7}56.9          \\ \cline{3-10} 
                       &                              & ROS                 & RF+ROS+NoC1   & \cellcolor[HTML]{96D3A7}95.3          & \cellcolor[HTML]{E8F4EE}18.6          & \cellcolor[HTML]{FBE9EC}27.6          & \cellcolor[HTML]{FCFCFF}22.1          & \cellcolor[HTML]{9CD5AC}79.2          & \cellcolor[HTML]{63BE7B}\textbf{64.6} \\ \cline{3-10} 
                       & \multirow{-4}{*}{No cluster} & Class   weights     & RF+CW+NoC1    & \cellcolor[HTML]{7BC890}95.4          & \cellcolor[HTML]{63BE7B}\textbf{20.6} & \cellcolor[HTML]{FBF8FB}30.4          & \cellcolor[HTML]{63BE7B}\textbf{24.4} & \cellcolor[HTML]{C4E5CF}77.1          & \cellcolor[HTML]{95D3A7}62.5          \\ \cline{2-10} 
                       &                              & No resampling       & RF+NoR+KM2    & \cellcolor[HTML]{B9E1C5}95.1          & \cellcolor[HTML]{D6EDDE}18.9          & \cellcolor[HTML]{FBF9FC}30.5          & \cellcolor[HTML]{B3DFC0}23.2          & \cellcolor[HTML]{8FD0A1}79.8          & \cellcolor[HTML]{9AD4AA}62.3          \\ \cline{3-10} 
                       &                              & RUS                 & RF+RUS+KM2    & \cellcolor[HTML]{C6E6D0}95.0          & \cellcolor[HTML]{B3DFC0}19.4          & \cellcolor[HTML]{B7E0C3}32.5          & \cellcolor[HTML]{73C589}24.2          & \cellcolor[HTML]{FBF7FA}73.8          & \cellcolor[HTML]{FBF2F5}57.6          \\ \cline{3-10} 
                       &                              & ROS                 & RF+ROS+KM2    & \cellcolor[HTML]{B4DFC1}95.1          & \cellcolor[HTML]{FCFCFF}18.3          & \cellcolor[HTML]{FBEFF2}28.7          & \cellcolor[HTML]{F4F9F8}22.2          & \cellcolor[HTML]{8ACE9D}80.1          & \cellcolor[HTML]{7BC890}63.6          \\ \cline{3-10} 
\multirow{-8}{*}{Tree} & \multirow{-4}{*}{clustering} & Class   weights     & RF+CW+NoC1    & \cellcolor[HTML]{7BC890}95.4          & \cellcolor[HTML]{63BE7B}20.6          & \cellcolor[HTML]{FBF8FB}30.4          & \cellcolor[HTML]{63BE7B}24.4          & \cellcolor[HTML]{C4E5CF}77.1          & \cellcolor[HTML]{95D3A7}62.5          \\ \hline
                       &                              & No   resampling     & ZIP+NoR+NoC1  & \cellcolor[HTML]{FBE8EB}94.4          & \cellcolor[HTML]{FAD7D9}14.6          & \cellcolor[HTML]{FBE5E8}26.8          & \cellcolor[HTML]{FBE1E4}18.9          & \cellcolor[HTML]{FBF9FC}74.0          & \cellcolor[HTML]{FBFBFE}58.0          \\ \cline{3-10} 
                       &                              & RUS                 & ZIP+RUS+NoC1  & \cellcolor[HTML]{FBDEE1}94.2          & \cellcolor[HTML]{FACFD2}13.9          & \cellcolor[HTML]{FBE1E4}26.1          & \cellcolor[HTML]{FBDBDE}18.1          & \cellcolor[HTML]{F8696B}61.1          & \cellcolor[HTML]{F8696B}50.6          \\ \cline{3-10} 
                       & \multirow{-3}{*}{No cluster} & ROS                 & ZIP+ROS+NoC1  & \cellcolor[HTML]{FBDADC}94.2          & \cellcolor[HTML]{FACFD2}13.9          & \cellcolor[HTML]{FBE4E7}26.7          & \cellcolor[HTML]{FBDCDF}18.2          & \cellcolor[HTML]{F8696B}61.2          & \cellcolor[HTML]{F8696B}50.6          \\ \cline{2-10} 
                       &                              & No resampling       & ZIP+NoR+KM2   & \cellcolor[HTML]{F9979A}93.1          & \cellcolor[HTML]{FAC7CA}13.1          & \cellcolor[HTML]{D3ECDC}31.9          & \cellcolor[HTML]{FBDEE1}18.5          & \cellcolor[HTML]{BAE2C6}77.6          & \cellcolor[HTML]{A5D9B4}61.8          \\ \cline{3-10} 
                       &                              & RUS                 & ZIP+RUS+KM2   & \cellcolor[HTML]{F99597}93.0          & \cellcolor[HTML]{FAC3C5}12.7          & \cellcolor[HTML]{FBFBFE}30.8          & \cellcolor[HTML]{FBD9DC}17.8          & \cellcolor[HTML]{FCFCFF}74.2          & \cellcolor[HTML]{FBE8EB}57.1          \\ \cline{3-10} 
\multirow{-6}{*}{ZIP}  & \multirow{-3}{*}{clustering} & ROS                 & ZIP+ROS+KM2   & \cellcolor[HTML]{F99799}93.0          & \cellcolor[HTML]{FAC4C7}12.8          & \cellcolor[HTML]{FCFCFF}30.9          & \cellcolor[HTML]{FBDADD}18.0          & \cellcolor[HTML]{FBFCFF}74.3          & \cellcolor[HTML]{FBE6E9}57.0          \\ \hline
\end{tabular}
}
\label{table:performance_evaluation_metrics}
\end{table}

\begin{figure*}  % spans both columns
\captionsetup{font=small}
\begin{subfigure}{0.32\textwidth}
\includegraphics[width=\linewidth]{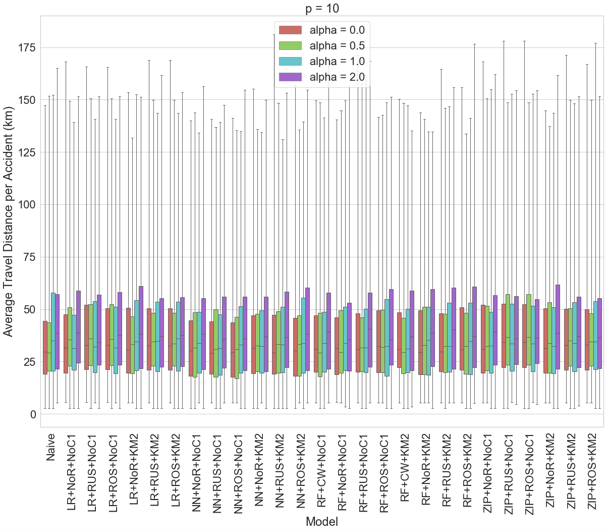}
 
\caption{$p=10$}
\end{subfigure}
\hfill % maximize the horizontal distance between the graphs
\begin{subfigure}{0.32\textwidth}
\includegraphics[width=\linewidth]{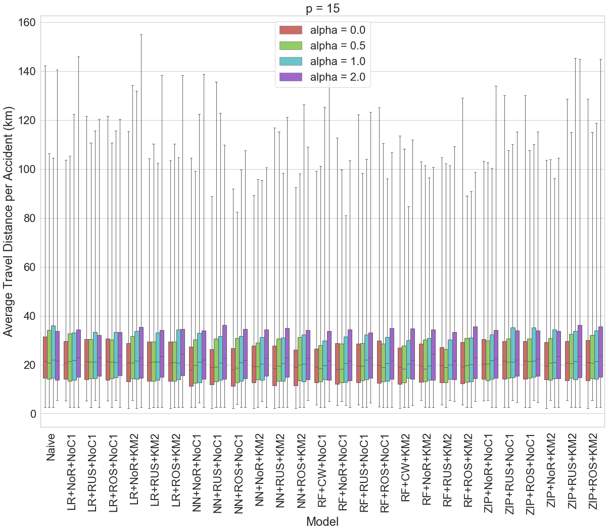}
 
\caption{$p=15$}
\end{subfigure}
\begin{subfigure}{0.33\textwidth}
\includegraphics[width=\linewidth]{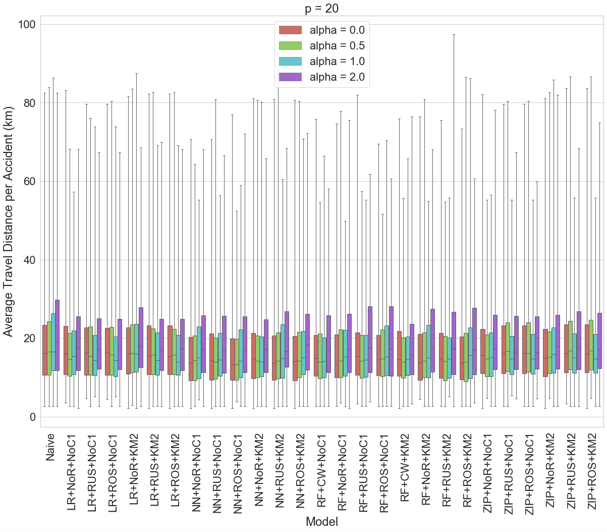}
 
\caption{$p=20$}
\end{subfigure}
% \bigskip  % some extra vertical whitespace
% \begin{subfigure}{0.33\textwidth}
% \includegraphics[width=\linewidth]{d.pdf}
% \caption{}
% \end{subfigure}
% \hfill % maximize the horizontal distance between the graphs
% \begin{subfigure}{0.33\textwidth}
% \includegraphics[width=\linewidth]{e.pdf}
% \caption{}
% \end{subfigure}
% \begin{subfigure}{0.33\textwidth}
% \includegraphics[width=\linewidth]{f.pdf}
% \caption{}
% \end{subfigure}

\caption{Total travel distance of responders per accident (a) 10 responders   (b) 15 responders, and (c) 20 responders; for example, for the naive model,  when $p=10$ and $\alpha=0$, the min., median, and max. of the average travel distance per accident is 2.65 km, 29.57 km, 147.26 km.} % Overall figure caption
\label{fig:complete_allocation}
\end{figure*}

\subsection{Forecasting}
We first evaluate the performance of the forecasting pipeline. We refer to the following abbreviations for brevity: LR (logistic regression), NN (neural networks), RF (random forests), ZIP (zero-inflated Poisson), RUS (random under sampling), ROS (random over sampling), NoC1 (No clustering) and KM2 (k-means clustering). Our baseline is based on the actual forecasting model that aids first responders in Tennessee. We refer to the baseline as the \textit{naive} model. The naive model essentially creates an empirical distribution based on historical incident data. Then, given a segment, a specific point in time, and the set of covariates induced by them, a realization of incident occurrence is sampled from the empirical distribution conditional on the covariates. We present results for each of the approaches in Table~\ref{table:performance_evaluation_metrics}. To understand the role and efficacy of each component of the pipeline, we present results with and without synthetic resampling and clustering. 

We present the major observations first --- neural networks and random forests outperform the naive model, logistic regression, and the zero-inflated Poisson regression model. Also, based on Table~\ref{table:performance_evaluation_metrics}, we see that while the naive model is fairly accurate, its accuracy is based on under-predicting accidents, as shown by its poor F-1 score. Also, clustering (even in isolation) generally improves the F1-score and accuracy of the forecasting models (for each method, compare the two rows that denote no resampling). We further observe a similar trend with synthetic sampling, which even in isolation usually results in an improvement in accuracy as well as F1-score (for each method, see the set of rows that denote no clustering and compare the rows that show resampling). The efficacy of the combination of clustering and oversampling is somewhat unclear though. We observe that typically, the combination \textit{slightly} under-performs in comparison to using one of the two approaches. We present three major takeaways from this observation --- first, synthetic sampling and clustering enable forecasting in sparse datasets significantly more than approaches that do not use them. However, we recommend practitioners to carefully evaluate each component of the proposed incident prediction pipeline on unseen data (test set) before deployment. Second, count-based models (zero-inflated Poisson regression) do not perform as well as binary classification models on sparse data. Third, it is important to note that while the resulting F1 scores might seem low in comparison to approaches on other data-driven problems, we claim that the improvement is significant in the context of extremely sparse and inherently random incidents like road accidents. We show the validity of this claim by simulating allocation and dispatch to accidents.

\begin{table}[t]
\captionsetup{font=small}
\centering
\caption{Average and maximum number accidents in each 4-h window that responders are not able to immediately respond because all responders are busy (the performance in each column is color coded; green is the best and reed is the worst.) Also, since all values for $p=0$ are zero, it is not included in this table.}
\resizebox{0.49\textwidth}{!}{%
\begin{tabular}{|l|c|c|c|c|c|c|c|c|c|c|c|c|c|c|c|c|}
\hline
             & \multicolumn{8}{c|}{average number of   unattended accidents}                                                                                                                                                                                         & \multicolumn{8}{c|}{maximum number of unattended accidents}                                                                                                                                                                                               \\ \hline
$p$           & \multicolumn{4}{c|}{\textbf{10}}                                                                                          & \multicolumn{4}{c|}{\textbf{15}}                                                                                          & \multicolumn{4}{c|}{\textbf{10}}                                                                                              & \multicolumn{4}{c|}{\textbf{15}}                                                                                          \\ \hline
$\alpha$        & \textbf{0}                   & \textbf{0.5}                 & \textbf{1}                   & \textbf{2}                   & \textbf{0}                   & \textbf{0.5}                 & \textbf{1}                   & \textbf{2}                   & \textbf{0}                    & \textbf{0.5}                  & \textbf{1}                    & \textbf{2}                    & \textbf{0}                   & \textbf{0.5}                 & \textbf{1}                   & \textbf{2}                   \\ \hline
Naïve        & \cellcolor[HTML]{FBC2C4}0.54 & \cellcolor[HTML]{FBB8BA}0.49 & \cellcolor[HTML]{FA9B9D}0.48 & \cellcolor[HTML]{FAA4A7}0.46 & \cellcolor[HTML]{F8696B}0.02 & \cellcolor[HTML]{FBCBCE}0.01 & \cellcolor[HTML]{FAB3B5}0.01 & \cellcolor[HTML]{FAB3B5}0.01 & \cellcolor[HTML]{FAB3B5}15.00 & \cellcolor[HTML]{FBD8DA}14.00 & \cellcolor[HTML]{FA9B9D}14.00 & \cellcolor[HTML]{F8696B}16.00 & \cellcolor[HTML]{FAB3B5}2.00 & \cellcolor[HTML]{FBCCCE}1.00 & \cellcolor[HTML]{FAB3B5}1.00 & \cellcolor[HTML]{F8696B}2.00 \\ \hline
LR+NoR+NoC1  & \cellcolor[HTML]{FBCCCE}0.54 & \cellcolor[HTML]{FCDFE2}0.47 & \cellcolor[HTML]{E6F3EC}0.42 & \cellcolor[HTML]{EBF5F0}0.42 & \cellcolor[HTML]{63BE7B}0.00 & \cellcolor[HTML]{63BE7B}0.00 & \cellcolor[HTML]{F8696B}0.01 & \cellcolor[HTML]{FCFCFF}0.01 & \cellcolor[HTML]{F98E90}16.00 & \cellcolor[HTML]{FCFCFF}13.00 & \cellcolor[HTML]{FA9B9D}14.00 & \cellcolor[HTML]{FCFCFF}12.00 & \cellcolor[HTML]{63BE7B}0.00 & \cellcolor[HTML]{63BE7B}0.00 & \cellcolor[HTML]{FAB3B5}1.00 & \cellcolor[HTML]{FCFCFF}1.00 \\ \hline
LR+RUS+NoC1  & \cellcolor[HTML]{FAA4A7}0.56 & \cellcolor[HTML]{F98789}0.52 & \cellcolor[HTML]{FBC2C4}0.46 & \cellcolor[HTML]{FAA4A7}0.46 & \cellcolor[HTML]{63BE7B}0.00 & \cellcolor[HTML]{63BE7B}0.00 & \cellcolor[HTML]{FAB3B5}0.01 & \cellcolor[HTML]{63BE7B}0.00 & \cellcolor[HTML]{F8696B}17.00 & \cellcolor[HTML]{F8696B}17.00 & \cellcolor[HTML]{F8696B}15.00 & \cellcolor[HTML]{F98E90}15.00 & \cellcolor[HTML]{63BE7B}0.00 & \cellcolor[HTML]{63BE7B}0.00 & \cellcolor[HTML]{FAB3B5}1.00 & \cellcolor[HTML]{63BE7B}0.00 \\ \hline
LR+ROS+NoC1  & \cellcolor[HTML]{FAA4A7}0.56 & \cellcolor[HTML]{FA9193}0.51 & \cellcolor[HTML]{FBC2C4}0.46 & \cellcolor[HTML]{FAB3B5}0.45 & \cellcolor[HTML]{63BE7B}0.00 & \cellcolor[HTML]{63BE7B}0.00 & \cellcolor[HTML]{FAB3B5}0.01 & \cellcolor[HTML]{63BE7B}0.00 & \cellcolor[HTML]{F8696B}17.00 & \cellcolor[HTML]{F8696B}17.00 & \cellcolor[HTML]{F8696B}15.00 & \cellcolor[HTML]{FAB3B5}14.00 & \cellcolor[HTML]{63BE7B}0.00 & \cellcolor[HTML]{63BE7B}0.00 & \cellcolor[HTML]{FAB3B5}1.00 & \cellcolor[HTML]{63BE7B}0.00 \\ \hline
LR+NoR+KM2   & \cellcolor[HTML]{FBD5D8}0.53 & \cellcolor[HTML]{96D2A7}0.41 & \cellcolor[HTML]{FCE9EC}0.43 & \cellcolor[HTML]{FBD0D3}0.44 & \cellcolor[HTML]{FA9B9D}0.02 & \cellcolor[HTML]{FA9A9D}0.01 & \cellcolor[HTML]{F8696B}0.01 & \cellcolor[HTML]{FAB3B5}0.01 & \cellcolor[HTML]{F8696B}17.00 & \cellcolor[HTML]{C9E7D3}12.00 & \cellcolor[HTML]{FBCCCE}13.00 & \cellcolor[HTML]{F98E90}15.00 & \cellcolor[HTML]{F8696B}3.00 & \cellcolor[HTML]{FBCCCE}1.00 & \cellcolor[HTML]{FAB3B5}1.00 & \cellcolor[HTML]{F8696B}2.00 \\ \hline
LR+RUS+KM2   & \cellcolor[HTML]{FBC2C4}0.54 & \cellcolor[HTML]{FBC2C4}0.48 & \cellcolor[HTML]{FCFCFF}0.42 & \cellcolor[HTML]{A7D9B5}0.40 & \cellcolor[HTML]{FCFCFF}0.01 & \cellcolor[HTML]{63BE7B}0.00 & \cellcolor[HTML]{63BE7B}0.00 & \cellcolor[HTML]{FAB3B5}0.01 & \cellcolor[HTML]{FAB3B5}15.00 & \cellcolor[HTML]{FAB3B5}15.00 & \cellcolor[HTML]{AFDDBD}11.00 & \cellcolor[HTML]{FCFCFF}12.00 & \cellcolor[HTML]{FCFCFF}1.00 & \cellcolor[HTML]{63BE7B}0.00 & \cellcolor[HTML]{63BE7B}0.00 & \cellcolor[HTML]{FCFCFF}1.00 \\ \hline
LR+ROS+KM2   & \cellcolor[HTML]{FBC2C4}0.54 & \cellcolor[HTML]{FBC2C4}0.48 & \cellcolor[HTML]{FCFCFF}0.42 & \cellcolor[HTML]{DAEEE1}0.41 & \cellcolor[HTML]{FCFCFF}0.01 & \cellcolor[HTML]{63BE7B}0.00 & \cellcolor[HTML]{63BE7B}0.00 & \cellcolor[HTML]{FCFCFF}0.01 & \cellcolor[HTML]{FAB3B5}15.00 & \cellcolor[HTML]{FAB3B5}15.00 & \cellcolor[HTML]{AFDDBD}11.00 & \cellcolor[HTML]{F98E90}15.00 & \cellcolor[HTML]{FCFCFF}1.00 & \cellcolor[HTML]{63BE7B}0.00 & \cellcolor[HTML]{63BE7B}0.00 & \cellcolor[HTML]{FCFCFF}1.00 \\ \hline
NN+NoR+NoC1  & \cellcolor[HTML]{7AC78F}0.45 & \cellcolor[HTML]{89CD9C}0.40 & \cellcolor[HTML]{FCF3F6}0.43 & \cellcolor[HTML]{B8E0C4}0.40 & \cellcolor[HTML]{FCFCFF}0.01 & \cellcolor[HTML]{63BE7B}0.00 & \cellcolor[HTML]{FAB3B5}0.01 & \cellcolor[HTML]{FCFCFF}0.01 & \cellcolor[HTML]{63BE7B}12.00 & \cellcolor[HTML]{96D2A7}11.00 & \cellcolor[HTML]{AFDDBD}11.00 & \cellcolor[HTML]{63BE7B}11.00 & \cellcolor[HTML]{FCFCFF}1.00 & \cellcolor[HTML]{63BE7B}0.00 & \cellcolor[HTML]{FAB3B5}1.00 & \cellcolor[HTML]{FCFCFF}1.00 \\ \hline
NN+RUS+NoC1  & \cellcolor[HTML]{A9DAB7}0.47 & \cellcolor[HTML]{A2D7B2}0.41 & \cellcolor[HTML]{FCF3F6}0.43 & \cellcolor[HTML]{FBC2C4}0.45 & \cellcolor[HTML]{63BE7B}0.00 & \cellcolor[HTML]{FBCBCE}0.01 & \cellcolor[HTML]{F8696B}0.01 & \cellcolor[HTML]{FAB3B5}0.01 & \cellcolor[HTML]{63BE7B}12.00 & \cellcolor[HTML]{96D2A7}11.00 & \cellcolor[HTML]{AFDDBD}11.00 & \cellcolor[HTML]{FCFCFF}12.00 & \cellcolor[HTML]{63BE7B}0.00 & \cellcolor[HTML]{FBCCCE}1.00 & \cellcolor[HTML]{FAB3B5}1.00 & \cellcolor[HTML]{FCFCFF}1.00 \\ \hline
NN+ROS+NoC1  & \cellcolor[HTML]{92D1A3}0.46 & \cellcolor[HTML]{96D2A7}0.41 & \cellcolor[HTML]{E6F3EC}0.42 & \cellcolor[HTML]{FCDFE2}0.43 & \cellcolor[HTML]{63BE7B}0.00 & \cellcolor[HTML]{63BE7B}0.00 & \cellcolor[HTML]{63BE7B}0.00 & \cellcolor[HTML]{FCFCFF}0.01 & \cellcolor[HTML]{63BE7B}12.00 & \cellcolor[HTML]{96D2A7}11.00 & \cellcolor[HTML]{AFDDBD}11.00 & \cellcolor[HTML]{FBD8DA}13.00 & \cellcolor[HTML]{63BE7B}0.00 & \cellcolor[HTML]{63BE7B}0.00 & \cellcolor[HTML]{63BE7B}0.00 & \cellcolor[HTML]{FCFCFF}1.00 \\ \hline
NN+NoR+KM2   & \cellcolor[HTML]{63BE7B}0.44 & \cellcolor[HTML]{89CD9C}0.40 & \cellcolor[HTML]{FCFCFF}0.42 & \cellcolor[HTML]{EBF5F0}0.42 & \cellcolor[HTML]{63BE7B}0.00 & \cellcolor[HTML]{63BE7B}0.00 & \cellcolor[HTML]{63BE7B}0.00 & \cellcolor[HTML]{63BE7B}0.00 & \cellcolor[HTML]{FAB3B5}15.00 & \cellcolor[HTML]{C9E7D3}12.00 & \cellcolor[HTML]{FCFCFF}12.00 & \cellcolor[HTML]{FAB3B5}14.00 & \cellcolor[HTML]{63BE7B}0.00 & \cellcolor[HTML]{63BE7B}0.00 & \cellcolor[HTML]{63BE7B}0.00 & \cellcolor[HTML]{63BE7B}0.00 \\ \hline
NN+RUS+KM2   & \cellcolor[HTML]{C1E4CC}0.48 & \cellcolor[HTML]{FCFCFF}0.45 & \cellcolor[HTML]{FCFCFF}0.42 & \cellcolor[HTML]{FCFCFF}0.42 & \cellcolor[HTML]{FBCBCE}0.01 & \cellcolor[HTML]{FBCBCE}0.01 & \cellcolor[HTML]{F8696B}0.01 & \cellcolor[HTML]{F8696B}0.02 & \cellcolor[HTML]{63BE7B}12.00 & \cellcolor[HTML]{96D2A7}11.00 & \cellcolor[HTML]{AFDDBD}11.00 & \cellcolor[HTML]{FCFCFF}12.00 & \cellcolor[HTML]{FCFCFF}1.00 & \cellcolor[HTML]{FBCCCE}1.00 & \cellcolor[HTML]{FAB3B5}1.00 & \cellcolor[HTML]{F8696B}2.00 \\ \hline
NN+ROS+KM2   & \cellcolor[HTML]{B5DFC2}0.48 & \cellcolor[HTML]{A2D7B2}0.41 & \cellcolor[HTML]{FCDFE2}0.44 & \cellcolor[HTML]{C9E7D3}0.41 & \cellcolor[HTML]{63BE7B}0.00 & \cellcolor[HTML]{63BE7B}0.00 & \cellcolor[HTML]{63BE7B}0.00 & \cellcolor[HTML]{FCFCFF}0.01 & \cellcolor[HTML]{FCFCFF}13.00 & \cellcolor[HTML]{96D2A7}11.00 & \cellcolor[HTML]{FA9B9D}14.00 & \cellcolor[HTML]{63BE7B}11.00 & \cellcolor[HTML]{63BE7B}0.00 & \cellcolor[HTML]{63BE7B}0.00 & \cellcolor[HTML]{63BE7B}0.00 & \cellcolor[HTML]{FCFCFF}1.00 \\ \hline
RF+NoR+NoC1  & \cellcolor[HTML]{FCFCFF}0.51 & \cellcolor[HTML]{E2F1E9}0.44 & \cellcolor[HTML]{E6F3EC}0.42 & \cellcolor[HTML]{FCFCFF}0.42 & \cellcolor[HTML]{63BE7B}0.00 & \cellcolor[HTML]{63BE7B}0.00 & \cellcolor[HTML]{63BE7B}0.00 & \cellcolor[HTML]{F8696B}0.02 & \cellcolor[HTML]{FCFCFF}13.00 & \cellcolor[HTML]{C9E7D3}12.00 & \cellcolor[HTML]{FCFCFF}12.00 & \cellcolor[HTML]{63BE7B}11.00 & \cellcolor[HTML]{63BE7B}0.00 & \cellcolor[HTML]{63BE7B}0.00 & \cellcolor[HTML]{63BE7B}0.00 & \cellcolor[HTML]{FCFCFF}1.00 \\ \hline
RF+RUS+NoC1  & \cellcolor[HTML]{C1E4CC}0.48 & \cellcolor[HTML]{89CD9C}0.40 & \cellcolor[HTML]{63BE7B}0.38 & \cellcolor[HTML]{FCEEF1}0.43 & \cellcolor[HTML]{FCFCFF}0.01 & \cellcolor[HTML]{FBCBCE}0.01 & \cellcolor[HTML]{63BE7B}0.00 & \cellcolor[HTML]{F8696B}0.02 & \cellcolor[HTML]{FCFCFF}13.00 & \cellcolor[HTML]{C9E7D3}12.00 & \cellcolor[HTML]{FBCCCE}13.00 & \cellcolor[HTML]{FBD8DA}13.00 & \cellcolor[HTML]{FCFCFF}1.00 & \cellcolor[HTML]{FBCCCE}1.00 & \cellcolor[HTML]{63BE7B}0.00 & \cellcolor[HTML]{F8696B}2.00 \\ \hline
RF+ROS+NoC1  & \cellcolor[HTML]{FBD5D8}0.53 & \cellcolor[HTML]{FCE9EC}0.46 & \cellcolor[HTML]{FCDFE2}0.44 & \cellcolor[HTML]{EBF5F0}0.42 & \cellcolor[HTML]{FCFCFF}0.01 & \cellcolor[HTML]{FBCBCE}0.01 & \cellcolor[HTML]{63BE7B}0.00 & \cellcolor[HTML]{63BE7B}0.00 & \cellcolor[HTML]{F98E90}16.00 & \cellcolor[HTML]{FCFCFF}13.00 & \cellcolor[HTML]{AFDDBD}11.00 & \cellcolor[HTML]{FAB3B5}14.00 & \cellcolor[HTML]{FCFCFF}1.00 & \cellcolor[HTML]{FBCCCE}1.00 & \cellcolor[HTML]{63BE7B}0.00 & \cellcolor[HTML]{63BE7B}0.00 \\ \hline
RF+CW+NoC1   & \cellcolor[HTML]{86CC99}0.46 & \cellcolor[HTML]{A2D7B2}0.41 & \cellcolor[HTML]{8ECFA0}0.40 & \cellcolor[HTML]{DAEEE1}0.41 & \cellcolor[HTML]{FCFCFF}0.01 & \cellcolor[HTML]{FBCBCE}0.01 & \cellcolor[HTML]{63BE7B}0.00 & \cellcolor[HTML]{FCFCFF}0.01 & \cellcolor[HTML]{63BE7B}12.00 & \cellcolor[HTML]{96D2A7}11.00 & \cellcolor[HTML]{FCFCFF}12.00 & \cellcolor[HTML]{FCFCFF}12.00 & \cellcolor[HTML]{FCFCFF}1.00 & \cellcolor[HTML]{FBCCCE}1.00 & \cellcolor[HTML]{63BE7B}0.00 & \cellcolor[HTML]{FCFCFF}1.00 \\ \hline
RF+NoR+KM2   & \cellcolor[HTML]{D8EDE0}0.49 & \cellcolor[HTML]{BCE2C7}0.42 & \cellcolor[HTML]{E6F3EC}0.42 & \cellcolor[HTML]{FCDFE2}0.43 & \cellcolor[HTML]{FCFCFF}0.01 & \cellcolor[HTML]{63BE7B}0.00 & \cellcolor[HTML]{63BE7B}0.00 & \cellcolor[HTML]{FCFCFF}0.01 & \cellcolor[HTML]{63BE7B}12.00 & \cellcolor[HTML]{FCFCFF}13.00 & \cellcolor[HTML]{AFDDBD}11.00 & \cellcolor[HTML]{FCFCFF}12.00 & \cellcolor[HTML]{FCFCFF}1.00 & \cellcolor[HTML]{63BE7B}0.00 & \cellcolor[HTML]{63BE7B}0.00 & \cellcolor[HTML]{FCFCFF}1.00 \\ \hline
RF+RUS+KM2   & \cellcolor[HTML]{CCE8D6}0.49 & \cellcolor[HTML]{89CD9C}0.40 & \cellcolor[HTML]{63BE7B}0.38 & \cellcolor[HTML]{EBF5F0}0.42 & \cellcolor[HTML]{63BE7B}0.00 & \cellcolor[HTML]{63BE7B}0.00 & \cellcolor[HTML]{63BE7B}0.00 & \cellcolor[HTML]{63BE7B}0.00 & \cellcolor[HTML]{FCFCFF}13.00 & \cellcolor[HTML]{63BE7B}10.00 & \cellcolor[HTML]{FCFCFF}12.00 & \cellcolor[HTML]{FCFCFF}12.00 & \cellcolor[HTML]{63BE7B}0.00 & \cellcolor[HTML]{63BE7B}0.00 & \cellcolor[HTML]{63BE7B}0.00 & \cellcolor[HTML]{63BE7B}0.00 \\ \hline
RF+ROS+KM2   & \cellcolor[HTML]{FCFCFF}0.51 & \cellcolor[HTML]{FCFCFF}0.45 & \cellcolor[HTML]{FCF3F6}0.43 & \cellcolor[HTML]{B8E0C4}0.40 & \cellcolor[HTML]{FBCBCE}0.01 & \cellcolor[HTML]{63BE7B}0.00 & \cellcolor[HTML]{FAB3B5}0.01 & \cellcolor[HTML]{FCFCFF}0.01 & \cellcolor[HTML]{FCFCFF}13.00 & \cellcolor[HTML]{FCFCFF}13.00 & \cellcolor[HTML]{AFDDBD}11.00 & \cellcolor[HTML]{FCFCFF}12.00 & \cellcolor[HTML]{FAB3B5}2.00 & \cellcolor[HTML]{63BE7B}0.00 & \cellcolor[HTML]{FAB3B5}1.00 & \cellcolor[HTML]{FCFCFF}1.00 \\ \hline
RF+CW+NoC1   & \cellcolor[HTML]{B5DFC2}0.48 & \cellcolor[HTML]{63BE7B}0.38 & \cellcolor[HTML]{63BE7B}0.38 & \cellcolor[HTML]{C9E7D3}0.41 & \cellcolor[HTML]{FCFCFF}0.01 & \cellcolor[HTML]{63BE7B}0.00 & \cellcolor[HTML]{63BE7B}0.00 & \cellcolor[HTML]{FCFCFF}0.01 & \cellcolor[HTML]{63BE7B}12.00 & \cellcolor[HTML]{63BE7B}10.00 & \cellcolor[HTML]{63BE7B}10.00 & \cellcolor[HTML]{FCFCFF}12.00 & \cellcolor[HTML]{FCFCFF}1.00 & \cellcolor[HTML]{63BE7B}0.00 & \cellcolor[HTML]{63BE7B}0.00 & \cellcolor[HTML]{FCFCFF}1.00 \\ \hline
ZIP+NoR+NoC1 & \cellcolor[HTML]{F0F7F4}0.51 & \cellcolor[HTML]{FCFCFF}0.45 & \cellcolor[HTML]{BAE1C6}0.41 & \cellcolor[HTML]{B8E0C4}0.40 & \cellcolor[HTML]{FA9B9D}0.02 & \cellcolor[HTML]{63BE7B}0.00 & \cellcolor[HTML]{63BE7B}0.00 & \cellcolor[HTML]{FCFCFF}0.01 & \cellcolor[HTML]{63BE7B}12.00 & \cellcolor[HTML]{FAB3B5}15.00 & \cellcolor[HTML]{FCFCFF}12.00 & \cellcolor[HTML]{FAB3B5}14.00 & \cellcolor[HTML]{F8696B}3.00 & \cellcolor[HTML]{63BE7B}0.00 & \cellcolor[HTML]{63BE7B}0.00 & \cellcolor[HTML]{FCFCFF}1.00 \\ \hline
ZIP+RUS+NoC1 & \cellcolor[HTML]{F8696B}0.59 & \cellcolor[HTML]{F8696B}0.53 & \cellcolor[HTML]{F8696B}0.51 & \cellcolor[HTML]{F8696B}0.48 & \cellcolor[HTML]{FBCBCE}0.01 & \cellcolor[HTML]{F8696B}0.02 & \cellcolor[HTML]{F8696B}0.01 & \cellcolor[HTML]{63BE7B}0.00 & \cellcolor[HTML]{F98E90}16.00 & \cellcolor[HTML]{F8696B}17.00 & \cellcolor[HTML]{F8696B}15.00 & \cellcolor[HTML]{F98E90}15.00 & \cellcolor[HTML]{FAB3B5}2.00 & \cellcolor[HTML]{F8696B}3.00 & \cellcolor[HTML]{F8696B}2.00 & \cellcolor[HTML]{63BE7B}0.00 \\ \hline
ZIP+ROS+NoC1 & \cellcolor[HTML]{F8696B}0.59 & \cellcolor[HTML]{F8696B}0.53 & \cellcolor[HTML]{F8696B}0.51 & \cellcolor[HTML]{F8696B}0.48 & \cellcolor[HTML]{FBCBCE}0.01 & \cellcolor[HTML]{F8696B}0.02 & \cellcolor[HTML]{F8696B}0.01 & \cellcolor[HTML]{63BE7B}0.00 & \cellcolor[HTML]{F98E90}16.00 & \cellcolor[HTML]{F8696B}17.00 & \cellcolor[HTML]{F8696B}15.00 & \cellcolor[HTML]{F98E90}15.00 & \cellcolor[HTML]{FAB3B5}2.00 & \cellcolor[HTML]{F8696B}3.00 & \cellcolor[HTML]{F8696B}2.00 & \cellcolor[HTML]{63BE7B}0.00 \\ \hline
ZIP+NoR+KM2  & \cellcolor[HTML]{A9DAB7}0.47 & \cellcolor[HTML]{EFF6F3}0.45 & \cellcolor[HTML]{FCFCFF}0.42 & \cellcolor[HTML]{63BE7B}0.37 & \cellcolor[HTML]{FA9B9D}0.02 & \cellcolor[HTML]{63BE7B}0.00 & \cellcolor[HTML]{FAB3B5}0.01 & \cellcolor[HTML]{FCFCFF}0.01 & \cellcolor[HTML]{FCFCFF}13.00 & \cellcolor[HTML]{FAB3B5}15.00 & \cellcolor[HTML]{FCFCFF}12.00 & \cellcolor[HTML]{FCFCFF}12.00 & \cellcolor[HTML]{F8696B}3.00 & \cellcolor[HTML]{63BE7B}0.00 & \cellcolor[HTML]{FAB3B5}1.00 & \cellcolor[HTML]{FCFCFF}1.00 \\ \hline
ZIP+RUS+KM2  & \cellcolor[HTML]{FAAEB1}0.55 & \cellcolor[HTML]{FBC2C4}0.48 & \cellcolor[HTML]{F97D7F}0.49 & \cellcolor[HTML]{FAA4A7}0.46 & \cellcolor[HTML]{63BE7B}0.00 & \cellcolor[HTML]{FBCBCE}0.01 & \cellcolor[HTML]{63BE7B}0.00 & \cellcolor[HTML]{FAB3B5}0.01 & \cellcolor[HTML]{F98E90}16.00 & \cellcolor[HTML]{FCFCFF}13.00 & \cellcolor[HTML]{F8696B}15.00 & \cellcolor[HTML]{FCFCFF}12.00 & \cellcolor[HTML]{63BE7B}0.00 & \cellcolor[HTML]{FBCCCE}1.00 & \cellcolor[HTML]{63BE7B}0.00 & \cellcolor[HTML]{FCFCFF}1.00 \\ \hline
ZIP+ROS+KM2  & \cellcolor[HTML]{FA9193}0.57 & \cellcolor[HTML]{FAAEB1}0.49 & \cellcolor[HTML]{FA9B9D}0.48 & \cellcolor[HTML]{FBC2C4}0.45 & \cellcolor[HTML]{63BE7B}0.00 & \cellcolor[HTML]{FBCBCE}0.01 & \cellcolor[HTML]{63BE7B}0.00 & \cellcolor[HTML]{FAB3B5}0.01 & \cellcolor[HTML]{F8696B}17.00 & \cellcolor[HTML]{FCFCFF}13.00 & \cellcolor[HTML]{F8696B}15.00 & \cellcolor[HTML]{FCFCFF}12.00 & \cellcolor[HTML]{63BE7B}0.00 & \cellcolor[HTML]{FBCCCE}1.00 & \cellcolor[HTML]{63BE7B}0.00 & \cellcolor[HTML]{FCFCFF}1.00 \\ \hline
\end{tabular}
}
\label{table:Distance}
\end{table}

\subsection{Allocation and Dispatch}

Our final goal is to enable our community partners save crucial response time to accidents. We now discuss how our forecasting models aid allocation and dispatch. We evaluate the entire combination of forecasting and dispatch on 186 temporal windows of 4 hours each, having a total of 1,865 incidents. We also vary the hyper-parameter $\alpha$ and the number of available responders $p$ (Tennessee currently has 10 responders, but we perform experiments on other realizations of $p$ nonetheless). We use two metrics to evaluate performance --- the average distance traveled by the responders and the number of incidents that cannot be attended due to unavailability of responders. We present the results in Fig.~\ref{fig:complete_allocation} and Table~\ref{table:Distance}.

\begin{figure}[h]
\captionsetup{font=small}
\centering
\includegraphics[width=\columnwidth]{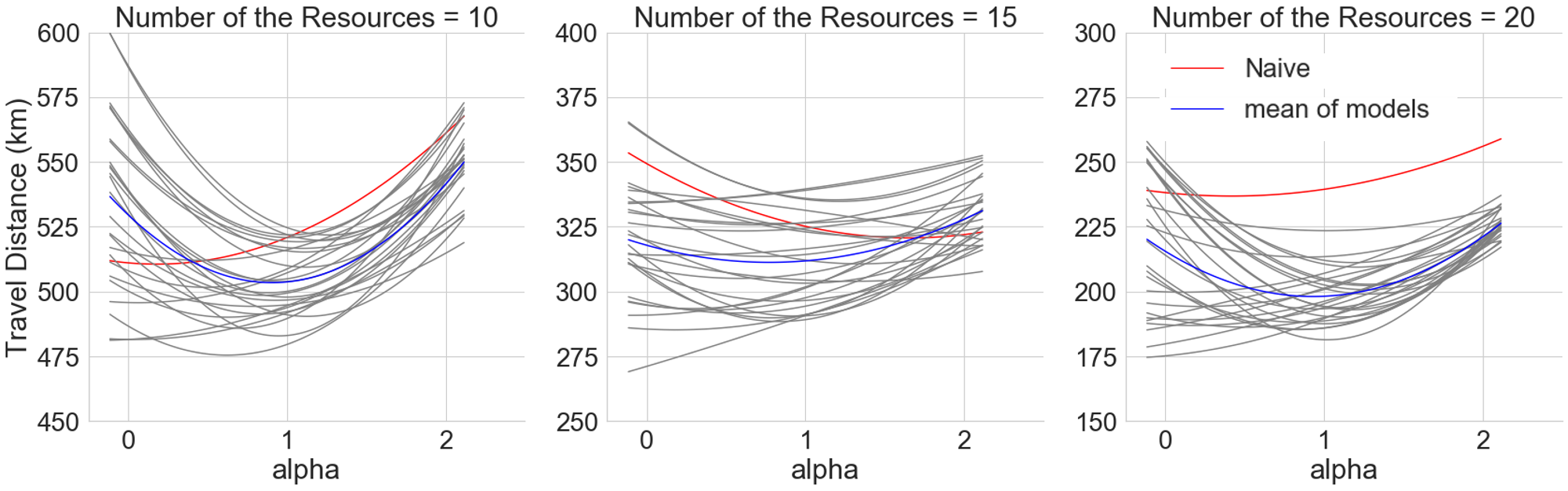}
\caption{Effect the of the hyper parameter $\alpha$ in allocation model on the performance of different models (the grey lines denote each learned model).}
\label{fig:alpha}
\end{figure}

An important observation is that our allocation approach, which adds a balancing term to the classical p-median problem, improves resource allocation in general; indeed, we observe this improvement in general across the spectrum of forecasting model used and the number of available responders. The maximum improvement we observe is a reduction of 3 km traveled by responders per incident (on average). This observation is particularly important for practitioners and first responders --- while it is important to allocate resources in areas with (relatively) high historical rates of occurrence, assigning a small number of responders to cover large areas can be detrimental to the overall goal of reducing response times. Intuitively, our approach penalizes additional burden on responders. However, we also note that a large penalty (value of $\alpha$) can result in increased response times. This is expected --- as $\alpha$ grows, it discourages the geographic spread of responders. Fig.\ref{fig:alpha} shows the influence of $\alpha$ on the performance of different models with a varying number of responders. Our empirical results show that $0.5 \leq \alpha \leq 1$ results in the optimal allocation of responders.

We also observe that our forecasting pipeline results in \textbf{significant} savings in response times (upto a maximum long-term reduction in response times by $10\%$) (see Figure~\ref{fig:complete_allocation}) and a reduction in the number of accidents that cannot be attended to (see Table~\ref{table:Distance}). It is important to understand the importance of this reduction. Prior work reports that a saving of only ten minutes of response time can reduce deaths due to road accidents by $33\%$~\cite{sanchez2010probability}. Allocating responders based on the random forest model provides the maximum reduction in response times (with neural networks being a close second). We provide three major takeaways based on our experiments on allocation and dispatch. First, forecasting models that provide the highest accuracy might not be the best candidates for allocation. This observation shows the importance of using a metric that focuses on false negatives and false positives (like the F-1 score) for sparse emergency incidents. Second, we point out that while traditional allocation models based on long-term (temporal) hotspots are widely used in the field, accurate short-term forecasting models can result in significant reduction in response times to accidents. Finally, leveraging the structure of the problem to improve classical resource allocation formulations can aid emergency response in the field.

\section{Conclusion}

Emergency response to incidents like road accidents is a major concern for first responders. Standard approaches to predicting road accident rarely scale to large geographic areas due to extremely high sparsity in data and difficulties in gathering data. In collaboration with the Tennessee Department of Transportation, we present a framework for forecasting extremely sparse spatial and temporal incidents like road accidents. We show how our approach for forecasting, based on a combination of non-spatial clustering, synthetic resampling, and learning from multiple data sources, outperforms forecasting methods used in the field. We also present a novel modification to a classical formulation for resource allocation. Through extensive simulations, we show how our pipeline results in significant reduction in response times to emergency incidents. Our implementation is open-source and can be used by other organizations that seek to optimize emergency response.

\section{Acknowledgement}
We acknowledge the support from the National Science Foundation (IIS-1814958), the Tennessee Department of Transportation (TDOT) for funding the research, and Google for providing computational resources.
\bibliographystyle{IEEEtran}
\bibliography{references}
\newpage
% \newpage
\appendix

We have made the code used in this study publicly available at the following link\footnote{\href{https://github.com/StatResp/KDD_IncidentPrediction}{https://github.com/StatResp/smart-comp\_IncidentPrediction}}. While detailed instructions on how to use the code can be found on the link, there are a few key files worth mentioning. 

\begin{itemize}
    \item \textbf{config/params.conf}: This file contains metadata and parameters that must be customized to a user's specific deployment. This includes configuration information such as file paths and dataset information. It also configures which models to run, model hyperparameters, the features to use in regression, clustering parameters, and the synthetic sampling to apply. More detail on the specific metadata can be found in the readme file.
    
    \item \textbf{run\_sample.py}: This is the main script for fitting and evaluating the forecasting models. Based on the metadata in the \textit{params.conf} file, this script loads and formats the data, calls sub-routines for clustering, synthetic sampling, tuning model specific hyper-parameters, and finally fit the desired models for each sliding test window. It then evaluates each model on a test set and outputs the following result files to the \textit{output/} directory: \textbf{DF\_Results.pkl} - pandas dataframe which contains the overall evaluation metrics (accuracy, etc.) for each of the models averaged over space and time, 
    %\textbf{DF\_Test\_metric.pkl} - dataframe containing the evaluation metrics for each model on each 4 hour time window (aggregated over space), 
    \textbf{DF\_Test\_spacetime.pkl} - dataframe containing the models' predictions for each 4-hour time window (used for resource allocation), \textbf{report.html} - html file which visualizes the evaluation results from the \textit{DF\_results file}. There is an example of the html results visualization provided in the repository. 
    
    \item \textbf{allocation/run\_allocation.py}: This script evaluates the models’ impact when integrated with an allocation model. It uses the output from \textit{run\_sample.py} as prediction inputs. Given a set of test incidents, the script performs allocation based on each model’s prediction output \\ (\textit{DF\_Test\_spacetime.pkl}), and then simulates dispatch to calculate the distance between incidents and their nearest responders. 
\end{itemize}

The data used in this study is proprietary, but we release a synthesized example dataset (in the \textit{sample\_data} folder) to demonstrate the expected data format. The data is formatted as a csv document; each row represents the features and incident counts for a 4 hour time window at a particular road segment. The specific feature names are detailed on the link provided.

\end{document}